\documentclass{aci}

\date{} 

\usepackage{enumitem}

\usepackage{url}

\usepackage{hyperref} 

\usepackage{txfonts}
\usepackage{cite}
\usepackage{mathabx}
\usepackage{subfigure}
\def\typeofarticle{Research article}
\def\currentvolume{vol}
\def\currentissue{n}

\def\ppages{1--15}
\def\DOI{DOI to be assigned}
\def\Received{Date Received}
\def\Accepted{Date Accepted}
\def\Published{Date Published}
\begin{document}

\onecolumn 

\begin{description}[labelindent=-0.5cm,leftmargin=3cm,style=multiline]

\item[\textbf{Citation}]{M. A. Shafiq, Z. Long, H. Di, and G. AlRegib, “A novel attention model for salient structure detection in seismic volumes,” Applied Computing and Intelligence, vol. 1, no. 1, pp. 31-45, Nov. 2021.}

\item[\textbf{DOI}]{\url{https://doi.org/10.3934/aci.2021002}}

\item[\textbf{Review}]{Date of publication: 22 November 2021}

\item[\textbf{Data and Codes}]{\url{https://ghassanalregib.info/}}

\item[\textbf{Bib}] {@article\{shafiq2021aci,\\
  title=\{A novel attention model for salient structure detection in seismic volumes\},\\
  author=\{Shafiq, M. and Long, Z. and Di, H. and AlRegib, G.\},\\
  journal=\{Applied Computing and Intelligence\},\\
  volume=\{1\},\\
  number=\{1\},\\
  pages=\{31--45\},\\
  year=\{2021\},\\
  month=\{Nov.\},\\
  publisher=\{AIMS Press\}\}
} 


\item[\textbf{Copyright}]{\textcopyright 2021 the Author(s), licensee AIMS Press. This is an open access article distributed under the terms of the Creative Commons Attribution License
(http://creativecommons.org/licenses/by/4.0).}

\item[\textbf{Contact}]{\href{mailto:long\_z@mercer.edu}{long\_z@mercer.edu}  OR \href{mailto:alregib@gatech.edu}{alregib@gatech.edu}\\ \url{https://ghassanalregib.info/} \\ }
\end{description}

\thispagestyle{empty}
\newpage
\clearpage
\setcounter{page}{1}

\title{A novel attention model for salient structure detection in seismic volumes}

\author{
  Muhammad Amir Shafiq\affil{1},
  Zhiling Long\affil{2,}\corrauth,
  Haibin Di\affil{3} 
  and Ghassan AlRegib\affil{4,}\corrauth
\footnote{M. Shafiq, Z. Long, and H. Di were affiliated with the OLIVES Lab, directed by Prof. Ghassan AlRegib, in the School of Electrical and Computer Engineering of Georgia Institute of Technology (https://ghassanalregib.info/).}
}

\shortauthors{the Author(s)}

\address{
  \addr{\affilnum{1}}{Apple Inc., Cupertino, CA, USA}
  \addr{\affilnum{2}}{Mercer University, Atlanta, GA, USA}
  \addr{\affilnum{3}}{Schlumberger Ltd., Houston, TX, USA}
  \addr{\affilnum{4}}{Georgia Institute of Technology, Atlanta, GA, USA}}
\corraddr{Email: long\_z@mercer.edu; alregib@gatech.edu.}

\editor{Name of Editor}

\begin{abstract}
A new approach to seismic interpretation is proposed to leverage visual perception and human visual system modeling. Specifically, a saliency detection algorithm based on a novel attention model is proposed for identifying subsurface structures within seismic data volumes. The algorithm employs 3D-FFT and a multi-dimensional spectral projection, which decomposes local spectra into three distinct components, each depicting variations along different dimensions of the data. Subsequently, a novel directional center-surround attention model is proposed to incorporate directional comparisons around each voxel for saliency detection within each projected dimension. Next, the resulting saliency maps along each dimension are combined adaptively to yield a consolidated saliency map, which highlights various structures characterized by subtle variations and relative motion with respect to their neighboring sections. A priori information about the seismic data can be either embedded into the proposed attention model in the directional comparisons, or incorporated into the algorithm by specifying a template when combining saliency maps adaptively. Experimental results on two real seismic datasets from the North Sea, Netherlands and Great South Basin, New Zealand demonstrate the effectiveness of the proposed algorithm for detecting salient seismic structures of different natures and appearances in one shot, which differs significantly from traditional seismic interpretation algorithms. The results further demonstrate that the proposed method outperforms comparable state-of-the-art saliency detection algorithms for natural images and videos, which are inadequate for seismic imaging data.
\end{abstract}


\maketitle

\section{Introduction}
\label{sec:intro}
One of the important steps in the exploration of hydrocarbons and reservoirs of oil and gas is the detection of subsurface structures, such as faults and salt domes. These structures span over several square kilometers and form structural traps due to their distinct lithologic properties such as low permeability. The detection of subsurface structures is accomplished through seismic interpretation, a process in which a geophysicist explores the seismic imaging data, both manually through visual inspection and with the assistance of computer programs, to identify relevant structures. As summarized in a recent survey~\cite{alregib2018survey}, various computational algorithms for seismic interpretation have been developed over the years. They focus on either detection or tracking of subsurface structures such as faults, salt domes, gas chimneys, and channels. They typically rely on analyzing characteristics observed in the imaging data, such as correlations, changes in intensity or contrast, and texture patterns, to name a few. Since seismic imaging data is represented in the form of 2-dimensional sections or 3-dimensional volumes, and visual inspection is utilized in a traditional interpretation by human interpreters, image or video processing algorithms have been heavily adopted in computational methods designed to assist seismic interpretation, which is evident from the survey. However, saliency detection based on human visual attention, although being studied extensively in the past two decades in the image/video processing community, has rarely been applied to computational seismic interpretation.  

Saliency detection techniques aim to identify visually prominent areas or objects within images and videos, which are salient areas or objects that most likely draw the attention of the human visual system (HVS)~\cite{Borji2013}. When applied to an image (or a video), a saliency detection algorithm will calculate a saliency value for each pixel (or voxel), resulting in a saliency map, which maps each pixel (or voxel) in the image (or video) to its associated saliency value. Correspondingly, areas in the image (or video) with high saliency values are identified as salient areas. Saliency detection mimics the attention mechanism of HVS, which enables HVS to focus its processing resources on elements conveying the most important information in a complex surrounding environment. Such a mechanism is described computationally by an attention model. The center-surround processing is commonly accepted as a general framework for attention models~\cite{Borji2013}. According to the general concept of center-surround processing, the level of saliency at a certain location is determined by comparing that location against its surrounding neighbors in terms of certain features. The higher is the center-surround difference, the greater is the level of saliency. Based on the general center-surround model, a specific saliency detection algorithm may have its own unique attention model by customizing within the general framework. If an attention model utilizes only low-level features extracted from the observed scene (such as color, contrast, motion, etc.), it is a bottom-up model. If a model also incorporates prior knowledge specific to a targeted application, it is a top-down model. 

Saliency detection has been studied extensively in the context of natural images and videos~\cite{zhang2008sun, hou2007saliency, guo2010novel, achanta2008salient, fang2014video, Long2015, schauerte2012quaternion, seo2009static, kadir2001saliency, itti2009bayesian, li2014visual, borji2015salient}. For example, in~\cite{zhang2008sun}, saliency is determined using the self-information of visual features. In~\cite{hou2007saliency}, saliency is calculated using spectral residues obtained in the frequency domain. In~\cite{guo2010novel}, a quaternion representation combining intensity, color, and motion features is utilized for saliency detection. In~\cite{achanta2008salient}, low-level features such as color and luminance are combined with a frequency-tuned approach. In~\cite{fang2014video}, saliency in videos is detected by combining saliency in the space domain and saliency in the time domain using a statistical uncertainty measure. Saliency detection for videos was also studied in~\cite{Long2015}, where 3-dimensional fast Fourier transform (3D-FFT) is performed locally, with the spectrum decomposed into spatial and temporal components to yield spatial and temporal saliency separately. In~\cite{alshawi2018}, an unsupervised approach was proposed to assess the reliability of detected saliency. In the past several years, application of deep learning techniques to saliency detection for natural images and videos has also attracted increasing interest, for which a comprehensive review can be found in~\cite{borji2021}.

Given that seismic interpretation is traditionally accomplished by human interpreters through visual inspection, saliency detection algorithms, if designed appropriately to imitate a human interpreter's visual attention, should be able to identify in seismic imaging data geological structures of interest to an interpreter as salient areas. Such algorithms will provide a brand new perspective to examine seismic imaging data. They have a potential to become very useful computational tools to assist in seismic interpretation, because they are capable of detecting various salient structures at the same time, even though those structures may differ in nature and in appearance. This will be drastically different from, and more efficient than most currently available computational interpretation techniques, for which an algorithm is typically designed to only detect structures of similar nature or appearance.

\begin{figure*}[ht]
  \centering
  \includegraphics[width=14.0cm]{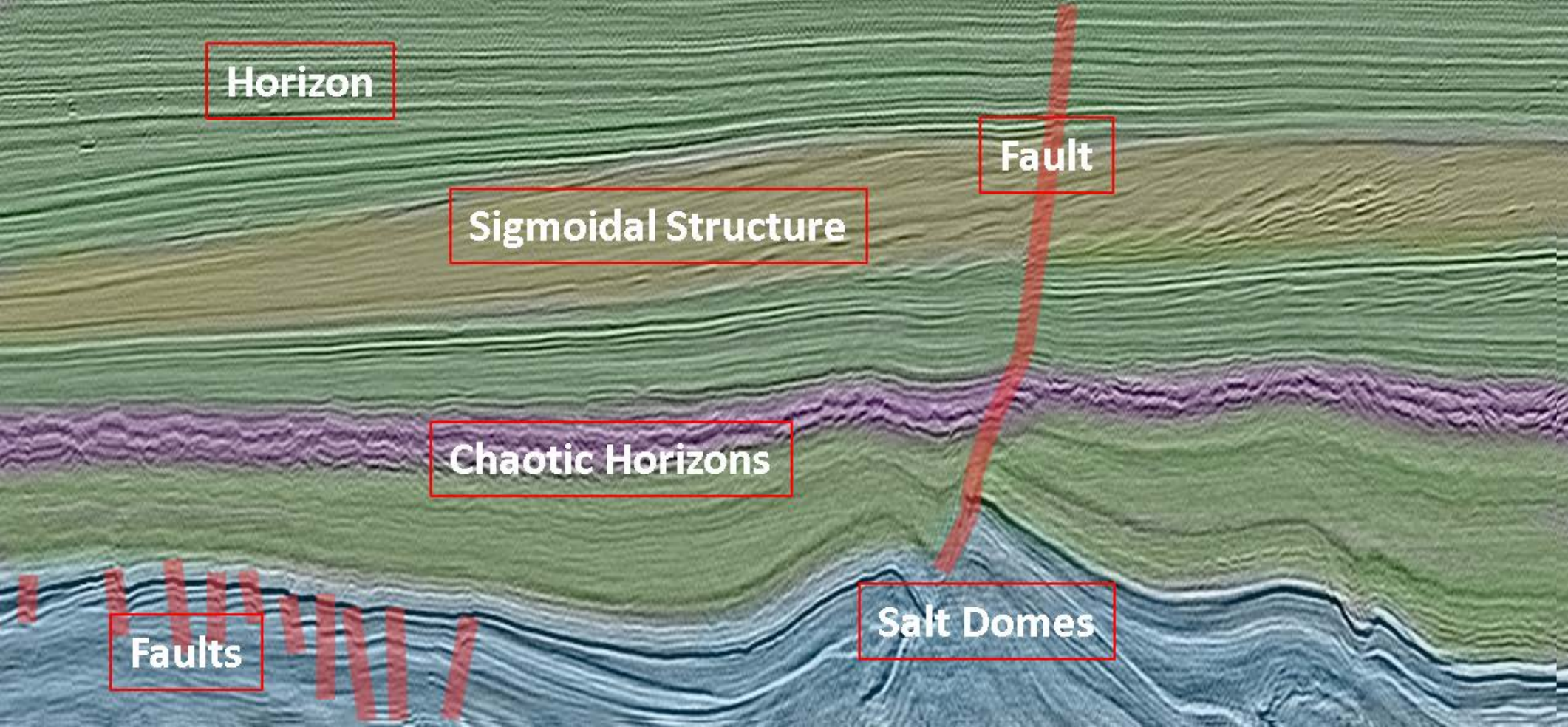}
  \caption{A typical seismic image (that is, a 2D cross section of a 3D seismic imaging data volume) with structures of interest for seismic interpretation~\cite{Long2018}. For illustration purpose, each structure has been annotated and colored distinctively.}\label{fig:Example_seismicImage}
\end{figure*}

In the past, there were rarely research activities reported on saliency detection for seismic imaging data. Among the very few pioneering explorations, an approach was proposed in~\cite{Drissi_Sal} for automated horizon picking based on salient features detection. In~\cite{ahujaexpert}, a study was carried out to gain heuristic knowledge of experts while interpreting seismic imaging data. Similarly, another study was reported in~\cite{sivarajah2014visual} of various saliency detection algorithms for natural images to observe which algorithm closely mimics the interpreter's visual attention during the interpretation of gravity and magnetic data, instead of seismic data. In our initial exploration on this topic, we presented two novel algorithms applying visual saliency to the detection and delineation of salt domes~\cite{Shafiq_ICASSP2016, ShafiqSalSaltDome}. These early studies applied saliency detection algorithms originally designed for natural images and videos directly to seismic imaging data. Although with some success for the specific tasks, such an approach is inadequate in general for seismic interpretation. Seismic imaging data is significantly different from natural images and videos, not only because the appearance is atypical, but also most structures are characterized by subtle changes in amplitude, texture, and contrast. For example, as illustrated in Fig.~\ref{fig:Example_seismicImage}, horizons are detected as continuous smooth horizontal lines. Chaotic horizons are similar to horizons, but with noisy fluctuations along the lines. A sigmoidal structure typically appears as S-shapped downlapping character extending outward and upward, with a fine texture. A salt dome is a mound of salt intruding upwards, typically appearing as a textured area enclosed by strong reflective boundaries. Finally, faults are identified when discontinuity is observed in any of the structures. Furthermore, the contextual information in the form of object's surroundings and geology also plays an important role in seismic interpretation. To detect salient structures in seismic imaging data, it is imperative to incorporate such domain knowledge as prior information into the attention model and the algorithm.

In this paper, we present a novel saliency detection algorithm for seismic imaging data in the form of 3D data volumes, which incorporates domain knowledge specific for seismic interpretation into the underlying attention model. Following our previous study reported in~\cite{Shafiq_ICASSP2016, ShafiqSalSaltDome}, we establish our new algorithm on the basis of the same 3D-FFT-based saliency detection method. The utilization of 3D-FFT helps capture energy variations within a seismic volume both effectively and computationally efficiently, and the multi-dimensional spectral projection tracks the variations along different dimensions reliably. Our major innovation in this new saliency detection algorithm is with the underlying attention model, in which we propose a unique directional center-surround comparison to replace the non-directional comparison commonly performed for natural images and videos. This directional operation is adjustable for any orientation, and aims to better accommodate seismic structures of directional nature, such as faults, gas chimneys, and horizons. Additionally, we design an adaptive scheme to automatically combine saliency values calculated from the multi-dimensional spectral projections into an overall saliency value. For evaluation, we use two entirely different seismic datasets under different geological settings to examine the effectiveness of the proposed saliency detection algorithm. We demonstrate that our innovative method outperforms comparable state-of-the-art algorithms in detecting salient subsurface structures of interest in seismic interpretation. We note that, although some of our preliminary results from this work were reported in~\cite{ICIP2017Shafiq, ICASSP2018Shafiq}, the algorithm we present in this paper has been improved by incorporating the adaptive scheme to obtain the final saliency map, and has been tested more extensively. 

\section{Proposed algorithm}
\begin{figure}[ht]
  \centering
  \includegraphics[width=16.0cm]{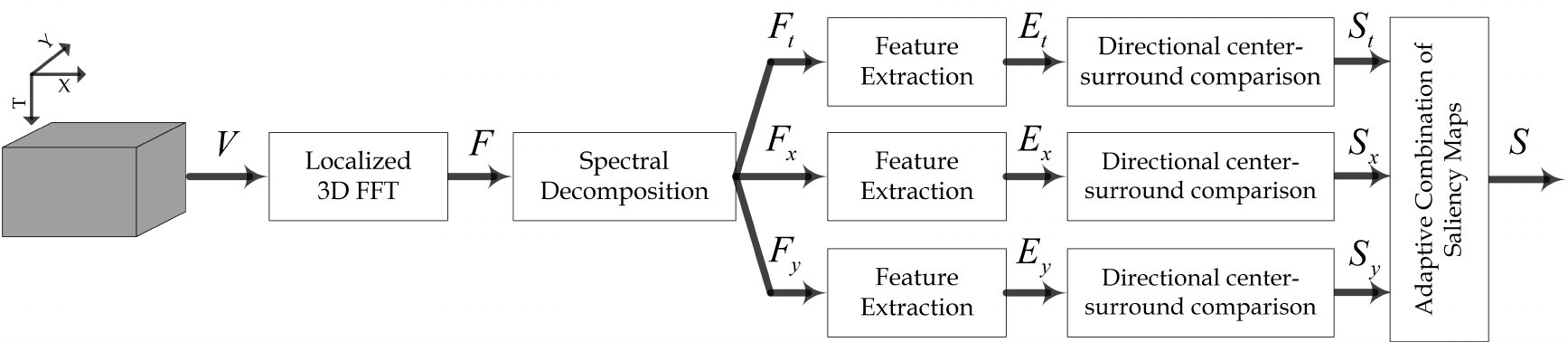}
  \caption{The block diagram of the proposed saliency detection algorithm.}\label{fig:Block_diag_Proposed}
\end{figure}

Our proposed saliency detection algorithm is illustrated in the block diagram shown in Fig.~\ref{fig:Block_diag_Proposed}. It starts with applying a localized 3D-FFT to the seismic volume being examined. Afterwards, the multi-dimensional spectral projection process decomposes a 3D-FFT spectrum to depict spatial variations along all three dimensions within the 3D volume. Subsequently, the application of directional center-surround (DCS) comparison highlights the salient variations along relevant orientations within the projected space. In the last step, the saliency maps generated from each projection are combined into the final saliency map using adaptive weights. In the following step-by-step detailed presentation of the algorithm, we assume a 3D seismic data volume $\boldsymbol V$ of size $T \times X \times Y$, where $T$ represents time (when reflection time is considered in imaging reconstruction) or depth (when actual subsurface depth is considered in imaging reconstruction), $X$ represents crosslines (direction perpendicular to the inlines), and $Y$ represents inlines (direction along which seismic imaging data was acquired); and we denote the computed saliency map as $\boldsymbol S$.

\begin{figure*}[!tb]
  \centering
    \subfigure[Spectral cube]{\includegraphics[width=.3\textwidth]{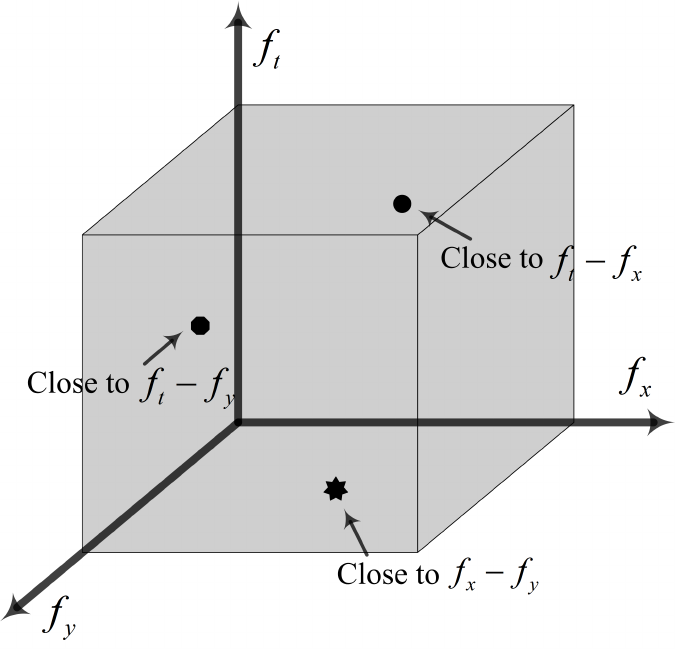}}
    \hspace{1cm}
    \subfigure[Plane projections]{\includegraphics[width=.53\textwidth]{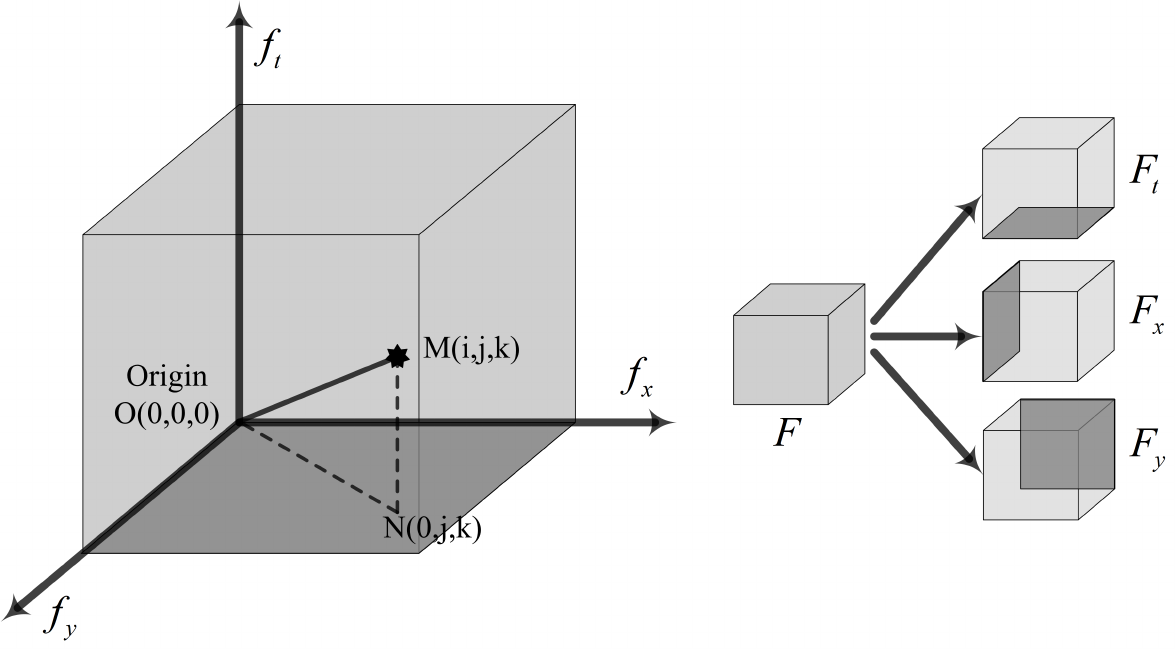}}
\caption{The illustration of spectral cube, plane projections, and decompositions.}
\label{fig:Spec_Decomp}
\end{figure*}

In the first step, we compute 3D-FFT of $\boldsymbol V$ using a local cube, with a sliding window having more than 50\% overlap, to yield a volume $\boldsymbol F$. In the second step, we perform decomposition of the spectral cube as explained in Fig.~\ref{fig:Spec_Decomp}. Within a 3D spectral cube in $f_t$-$f_x$-$f_y$ coordinate system, if a spectral point is closer to $f_x$-$f_y$-plane, then its projection on $f_x$-$f_y$-plane, i.e., along $f_t$-direction will depict variations more prominently as compared to the projections on $f_t$-$f_y$ or $f_t$-$f_x$ planes. Therefore, we decompose the 3D spectral cube by projecting the spectral point $F[i,j,k]$ along different dimensions as
\begin{equation}
    \boldsymbol{F}_m[i,j,k] = \boldsymbol{F}[i,j,k] \times \frac{\boldsymbol{P}_m[i,j,k]}{\sqrt{i^2+j^2+k^2}}, \quad m \in \{ t,x,y \},
\end{equation}
where projections $\boldsymbol{P}_m$ along time, crossline, and inline dimensions are represented by $\sqrt{j^2+k^2}$, $\sqrt{i^2+k^2}$, and $\sqrt{i^2+j^2}$ respectively. $\boldsymbol{F}$ represents the multi-dimensional FFT domain defined as $\boldsymbol{F} =  \mathcal{K} \oasterisk {C}_{n}$, where ${C}_{n}$ represents a local cube of side length $n$ within volume $\boldsymbol V$, $\oasterisk$ represents tensor product, and $\mathcal{K}$ is the Kronecker matrix defined as
\begin{equation}
\mathcal{K} = D_x \otimes D_y \otimes D_z,
\end{equation}
where $D_x$, $D_y$, and $D_z$ are transformation matrixes for the discrete Fourier transform (DFT), and $\otimes$ represents the Kronecker product. The spectral projections are not computed at the center of the local cube, as it represents the zero Hertz component and does not reflect any changes along the three planes. 

\begin{figure}[t]
  \centering
  \includegraphics[width=7.0cm]{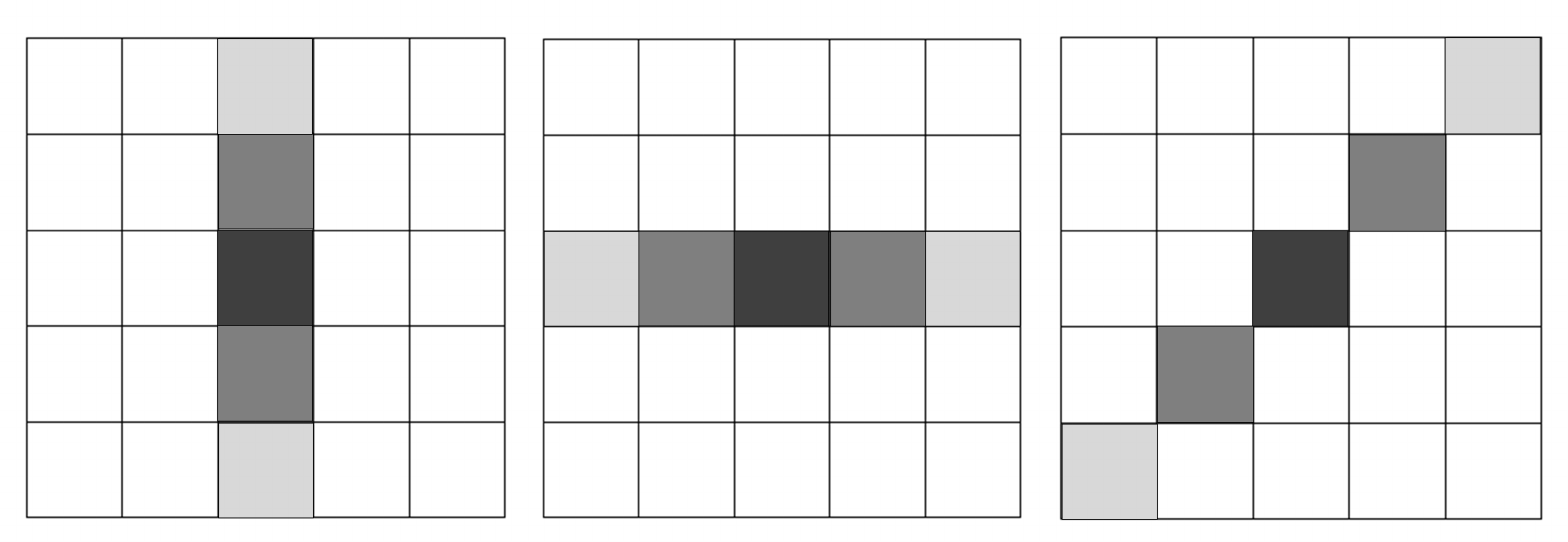}
  \caption{The illustration of DCS comparison along $t$, $x$, and $t$-$x$ directions, respectively.}\label{fig:Directional_M}
\end{figure}

In the third step, we extract features also known as spectral energies, $\boldsymbol E_m$, where $m \in \{t,x,y\}$, from spectral projections based on the absolute mean of local cube. The process of feature extraction enhances motion variations and provides pixel-level descriptions based on spectral energies in the neighborhood of a voxel. 

In the fourth step of the proposed method, we apply the DCS comparison to construct the saliency maps $\boldsymbol{S_m}$ using $\boldsymbol{E_m}$ as
\begin{align}\label{eqn:S_x}
\boldsymbol{S}_m[t,x,&y]  = \frac{1}{Q} \sum_{i_0,j_0,r_0} | \boldsymbol{E}_m[t,x,y] - \nonumber \\
& w \cdot \boldsymbol{E}_m[t+i_0,x+j_0,y+r_0] |,~~ m \in \{t,x,y\},
\end{align}
where $Q$ represents the total number of points included in the summation and $w$ represents Gaussian weights. $i_0$, $j_0$, $r_0$ are chosen such that point $[t+i_0,x+j_0,y+r_0]$ is in the immediate neighborhood of point $[t,x,y]$, such as within a directional window centered at $[t,x,y]$ as depicted in Fig.~\ref{fig:Directional_M}. In order to incorporate a priori information, we can either apply different directional filters pertaining to desired orientations, sizes, or shapes in DCS comparison, or we can apply different weights to various spectral projections to enhance any desired feature within seismic volume. Alternatively, we can also combine both to yield optimum results. DCS comparisons for selected orientations along $t$, $x$, and $t$-$x$ directions are illustrated in Fig.~\ref{fig:Directional_M}, where level of brightness indicates the magnitude of associated weights for each voxel, with darkest being the highest. Similarly, we can create templates to embed desired orientation, size, or shape information in DCS comparison. 

\begin{figure}[t]
  \centering
  \includegraphics[width=7.0cm]{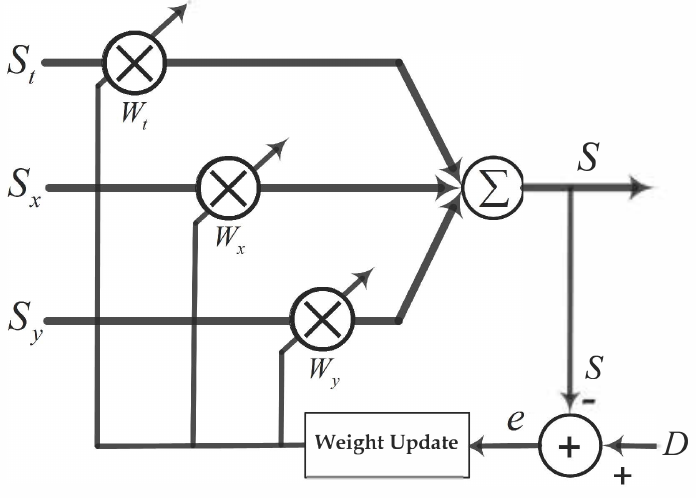}
  \caption{Adaptive wights update.}\label{fig:AdaptiveWtUpdate}
\end{figure}

Finally, the saliency map $\boldsymbol{S}$ is obtained via adaptive combination of directional saliency maps along each dimension using a block diagram shown in Fig.~\ref{fig:AdaptiveWtUpdate}. $\boldsymbol{D}$ represents a desired map which can be either manually labelled seismic volume that highlights desired structures within the volume, or a saliency map selected from the directional saliency maps $S_m, m \in \{t,x,y\}$. For weights update, we implemented least mean square (LMS), normalized least mean square (NLMS), and recursive least square (RLS) algorithms. The saliency map $\boldsymbol{S}$, which is of the same size as that of $\boldsymbol{V}$ is obtained by the dot product of weight vector and individual saliency maps as

\begin{equation} \label{eqn:Sadap}
\boldsymbol{S}= 
    \begin{bmatrix}
        W_t  &  W_x  &  W_y
    \end{bmatrix} 
\cdot 
    \begin{bmatrix}
        \boldsymbol{S}_t  &  \boldsymbol{S}_x & \boldsymbol{S}_y
    \end{bmatrix}^T
\end{equation}
where $W_m, m \in \{t,x,y\}$ represents adaptive weights obtained via LMS, NLMS, or RLS. Alternatively, the weights $W_m$ can be set either equally to construct a saliency map with equal distribution of projections along each axis or empirically to highlight certain features along any particular dimension. 

\begin{figure}[ht]
  \centering
  \includegraphics[width=.9\textwidth]{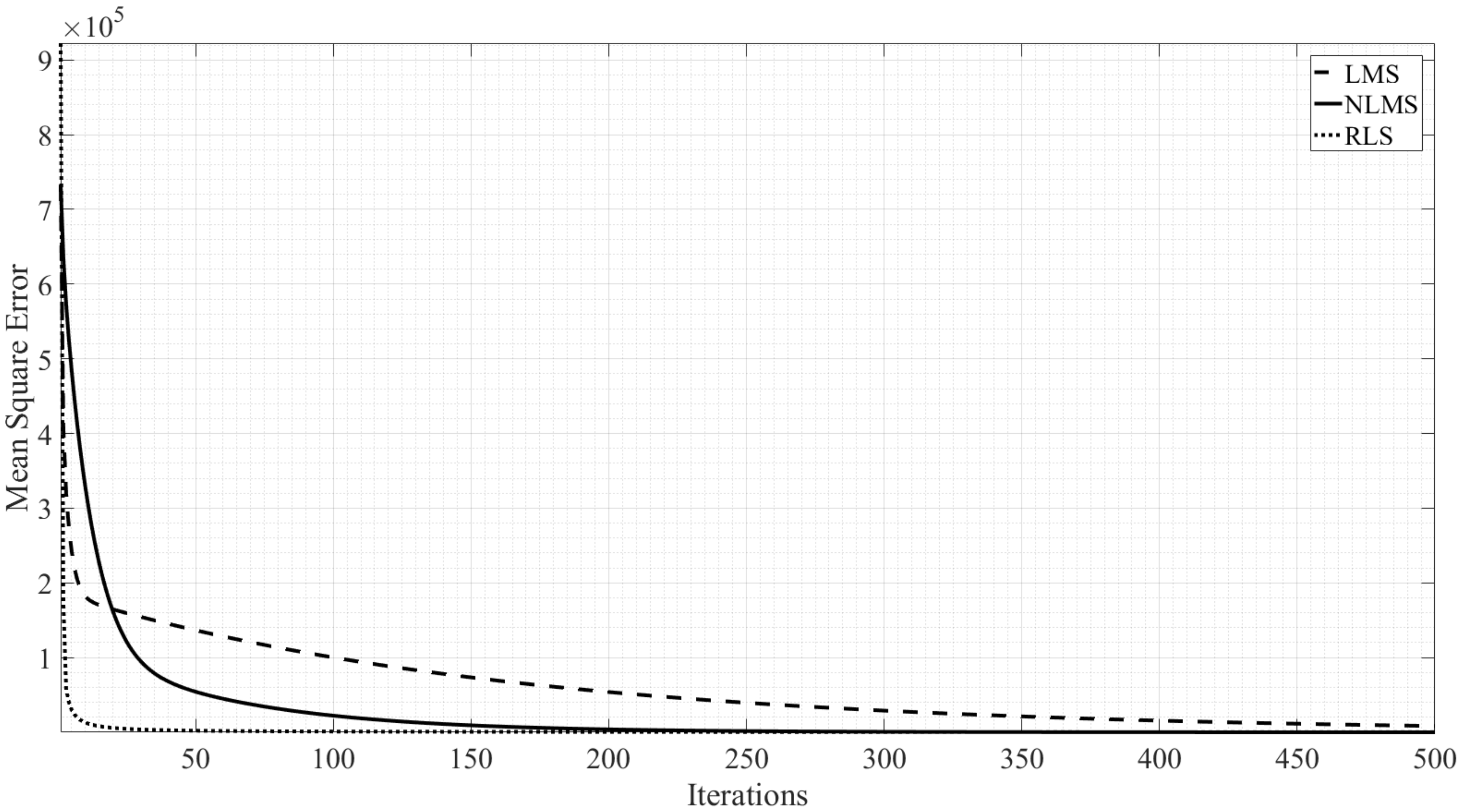}
  \caption{The mean square error of LMS, NLMS, and RLS algorithms for adapting a saliency map along time dimension.}\label{fig:MSE_GSB}
\end{figure}

The adaptive combination of individual saliency maps along each dimension provides a non-parametric way of computing $\boldsymbol{S}$ that can not only be tuned to highlight specific structures, but can also be adapted to any individual saliency map. The mean square error (MSE) of various adaptive algorithms for adapting weights to a certain saliency map is shown in Fig.~\ref{fig:MSE_GSB}. It can be observed that all algorithms converge to adapt the desired map, with RLS requiring the least number of iterations. However, RLS is computationally expensive because of matrix inversion; therefore, we have used NLMS for adaptive combination of individual saliency maps in this paper. Finally, the selection of desired map for weight adaption is left as an input by the interpreter. For example, as in the case of interpreting normal and reverse faults, which are usually oriented along time/depth dimension, an interpreter can specify a labelled seismic section for faults adaption that will automatically calculate the weights and yield an optimum saliency map without any manual tuning or empirical weight selection. It is worth mentioning that we do not have to adapt each seismic section for extracting desired features. The weights are adapted to any individual saliency map or given desired map only once for whole seismic volume and can be adapted again after every $N$ seismic sections, if required.

\section{Experimental results}
\label{sec:Results}
In this section, we present the results of the proposed saliency detection algorithm on two real seismic datasets acquired from the Netherlands offshore F3 block in the North Sea and Great South Basin (GSB), New Zealand, respectively. Both datasets consist of standard 3D seismic imaging data reconstructed from reflected vibration waves collected over an area of interest. The datasets are available in the standard SEG-Y format. Detailed information about the datasets can be found at~\cite{F3DataLink,GSBDataLink}. These datasets comprise of different subsurface structures that are characterized by subtle variations in intensity and texture, making them extremely challenging and difficult to detect if regular saliency detection algorithms for natural images and videos are used. 

We note that visual inspection is the dominant method of evaluation in seismic interpretation. Although attempts were made to propose quantitative methods of evaluation in recent years~\cite{Long2018,Shafiq_ICASSP2016,ShafiqSalSaltDome,Wang2014ICIP,Wang2015ICIP,Wang2015Geo,Long2015SEG}, such methods are limited in being tied to specific structures or tasks. They are still far from being widely accepted in the community. As such, we follow the common practice to evaluate our results in this paper. We show the saliency maps generated from our proposed algorithm side by side with those from comparable state-of-the-art saliency detection algorithms for a qualitative comparison. As a general rule, a well-designed seismic interpretation algorithm should not only resolve spatial variations along different directions within a seismic volume, but also highlight geological structures that are hardly conspicuous with respect to its surrounding environment. 

\subsection{F3 Block in the North Sea, Netherlands}
\begin{figure*}[!tb]
  \centering
    \subfigure[A typical seismic time slice]{\includegraphics[width=.425\textwidth]{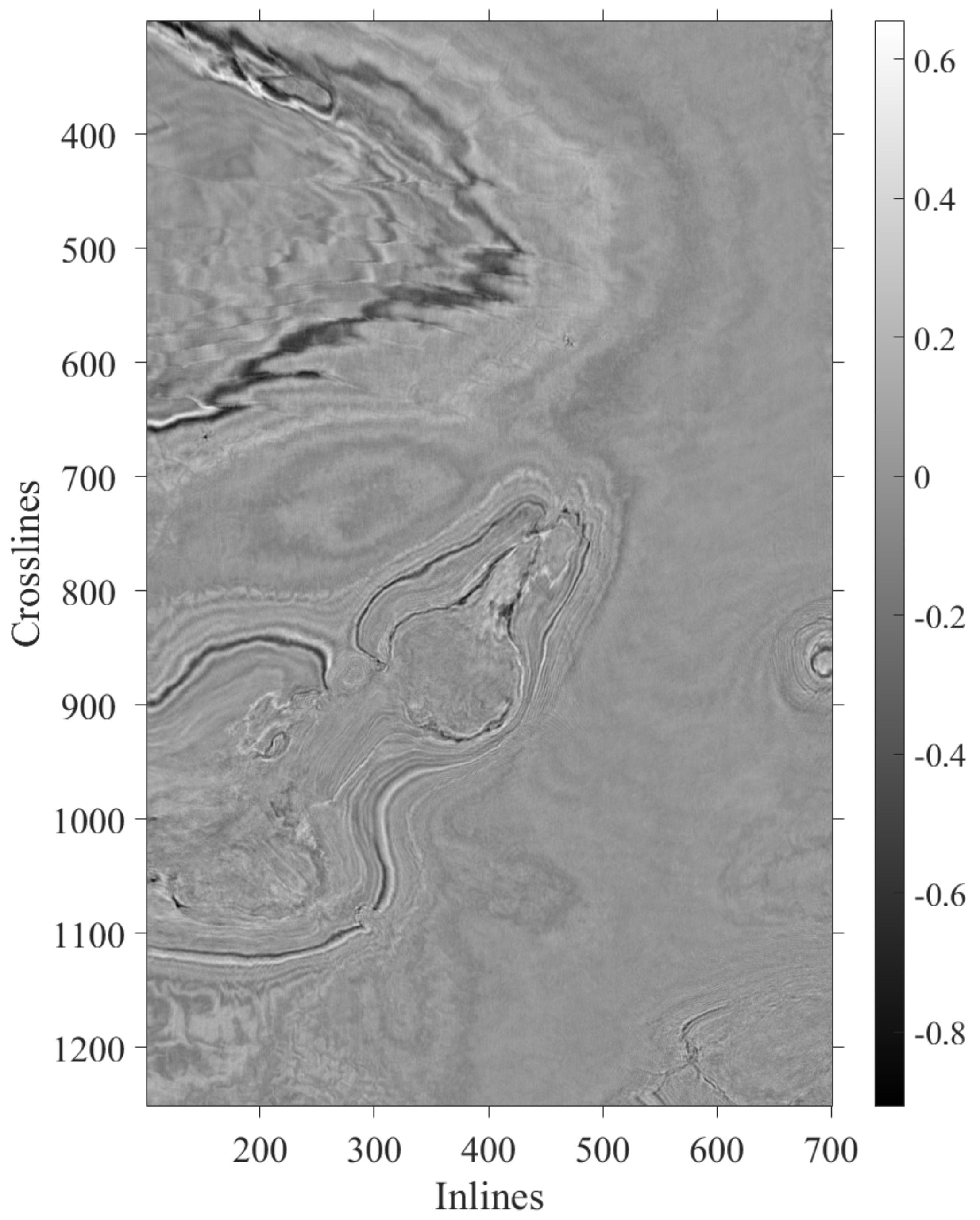}\label{fig:SS400_F3_Time_1600ms}}
    \hspace{1cm}
    \subfigure[Saliency map]{\includegraphics[width=.42\textwidth]{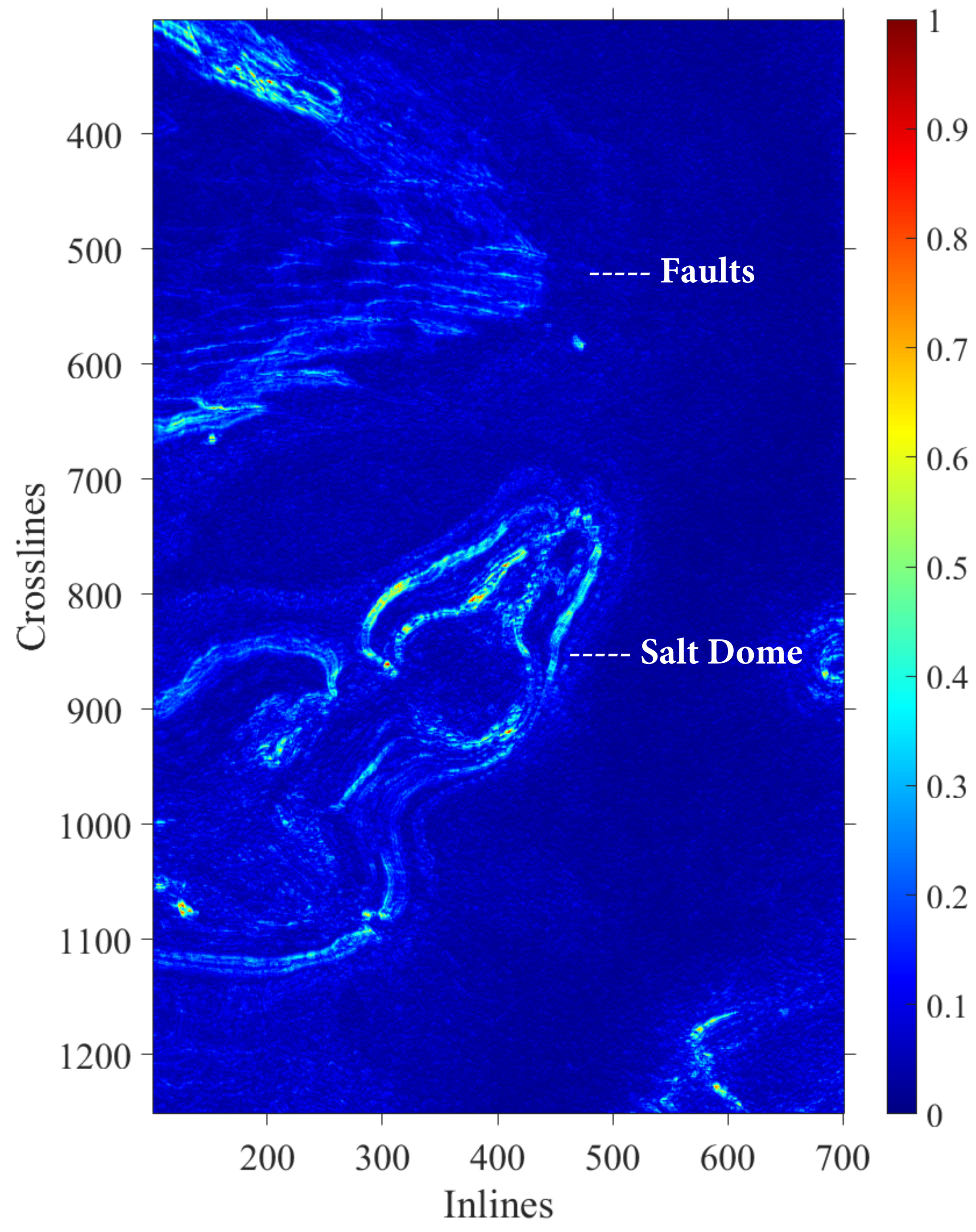}\label{fig:SS400_SalMap_F3_Time_1600ms}}
\caption{A seismic time section within F3 block at 1600ms and its saliency map generated using our proposed algorithm.}
\label{fig:TimeSection1600F3}
\end{figure*}

A typical seismic time section (i.e., a cross section of the 3D data volume along the t-direction, assuming an x-y-t coordinate system depicted in Fig.~\ref{fig:Block_diag_Proposed}) from the F3 block and its saliency map obtained using our proposed algorithm in Fig.~\ref{fig:Block_diag_Proposed} are shown in Fig.~\ref{fig:TimeSection1600F3}a-b, respectively. It can be observed from the saliency map that it effectively captures salt dome structures and, at the same time, pleasantly highlights multiple faults as shown in the top left of the saliency map. 

\begin{figure*}[!ht]
  \centering
  \subfigure[A typical seismic inline section.]{\includegraphics[width=.325\textwidth]{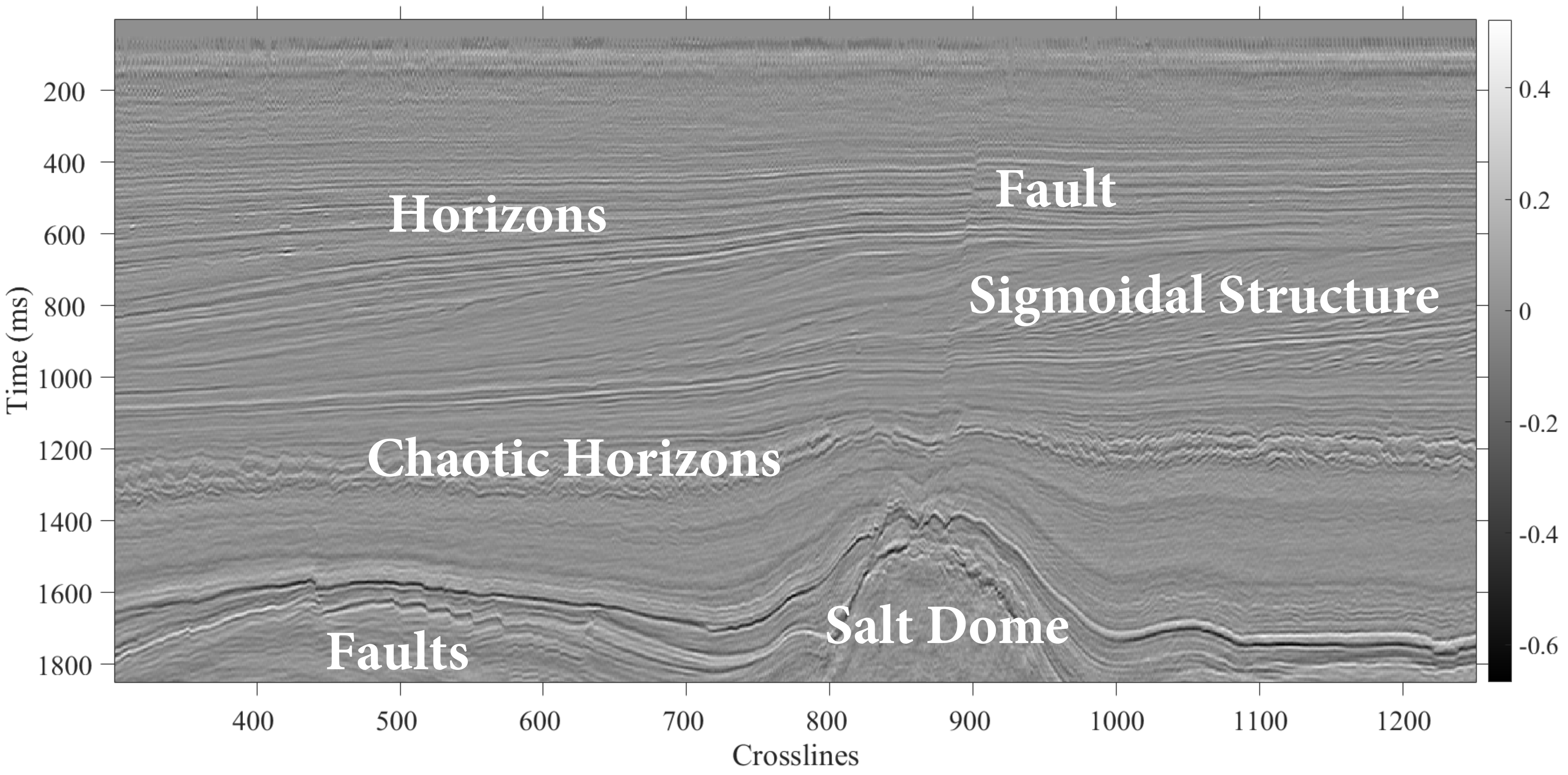}\label{fig:SS245F3}}
  \subfigure[SUN~\cite{zhang2008sun}]{\includegraphics[width=.325\textwidth]{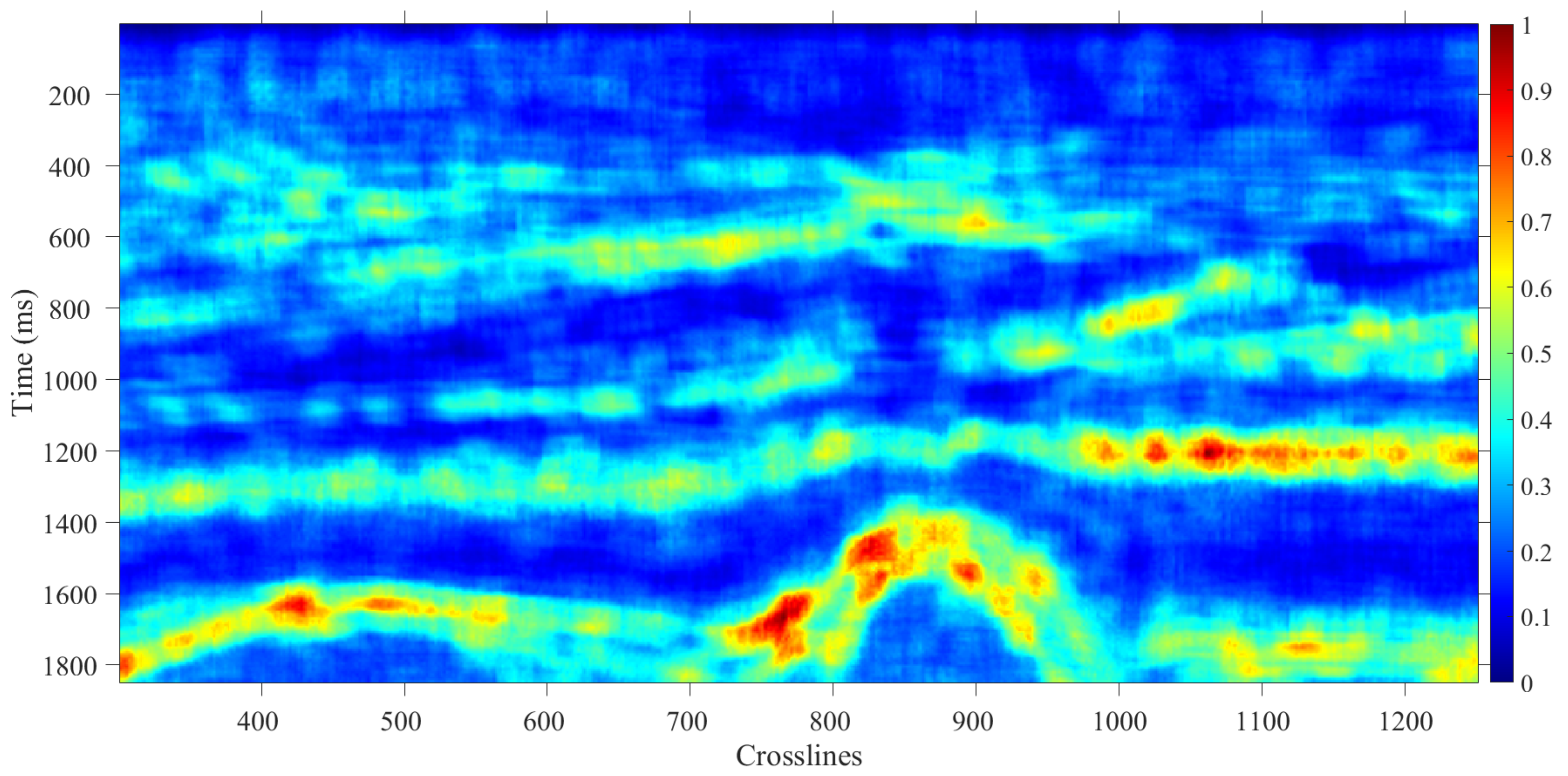}\label{fig:ISF3}}
  \subfigure[SR~\cite{hou2007saliency}]{\includegraphics[width=.325\textwidth]{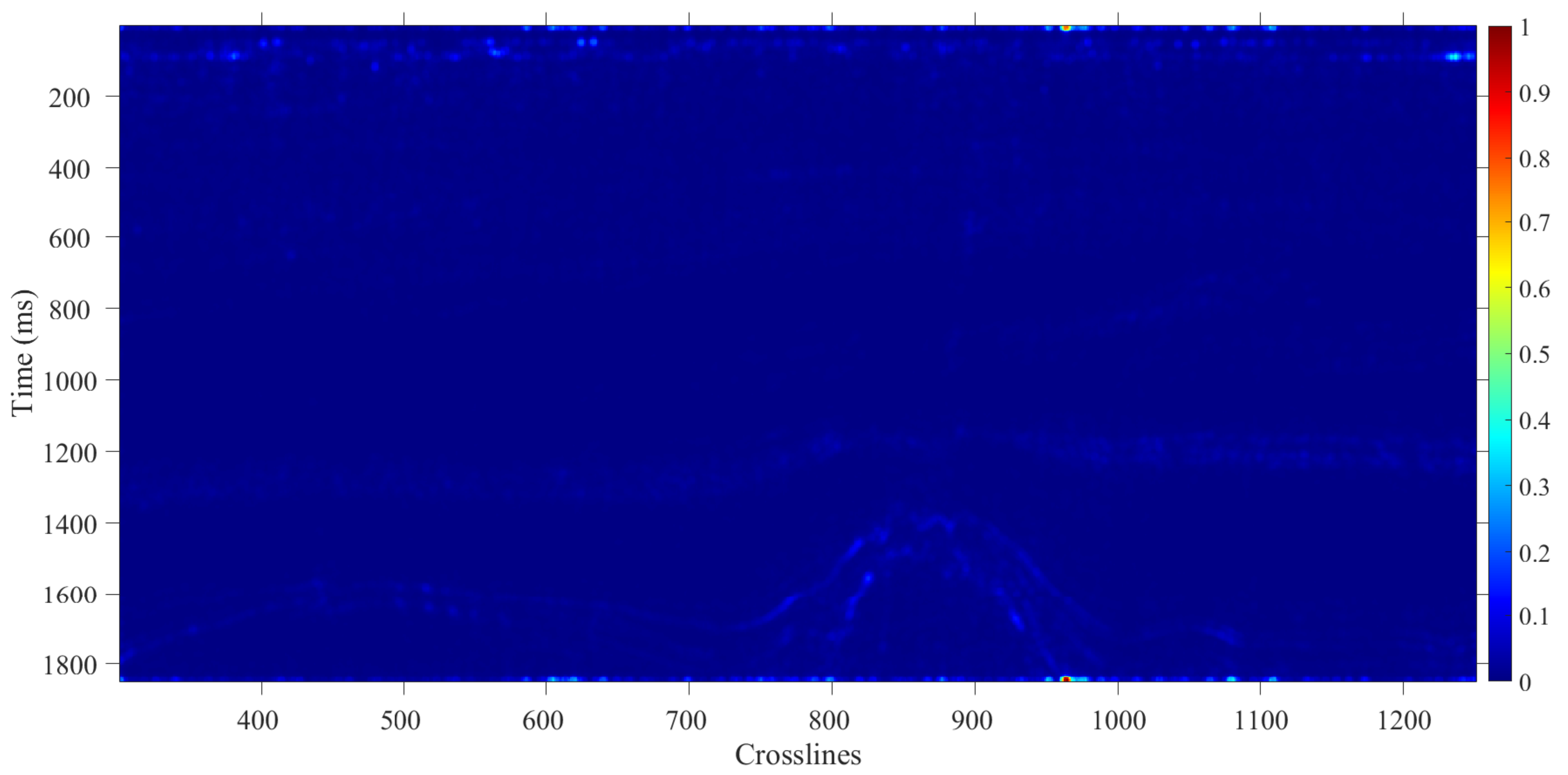}\label{fig:SResF3}}
  \subfigure[PQFT~\cite{guo2010novel}]{\includegraphics[width=.325\textwidth]{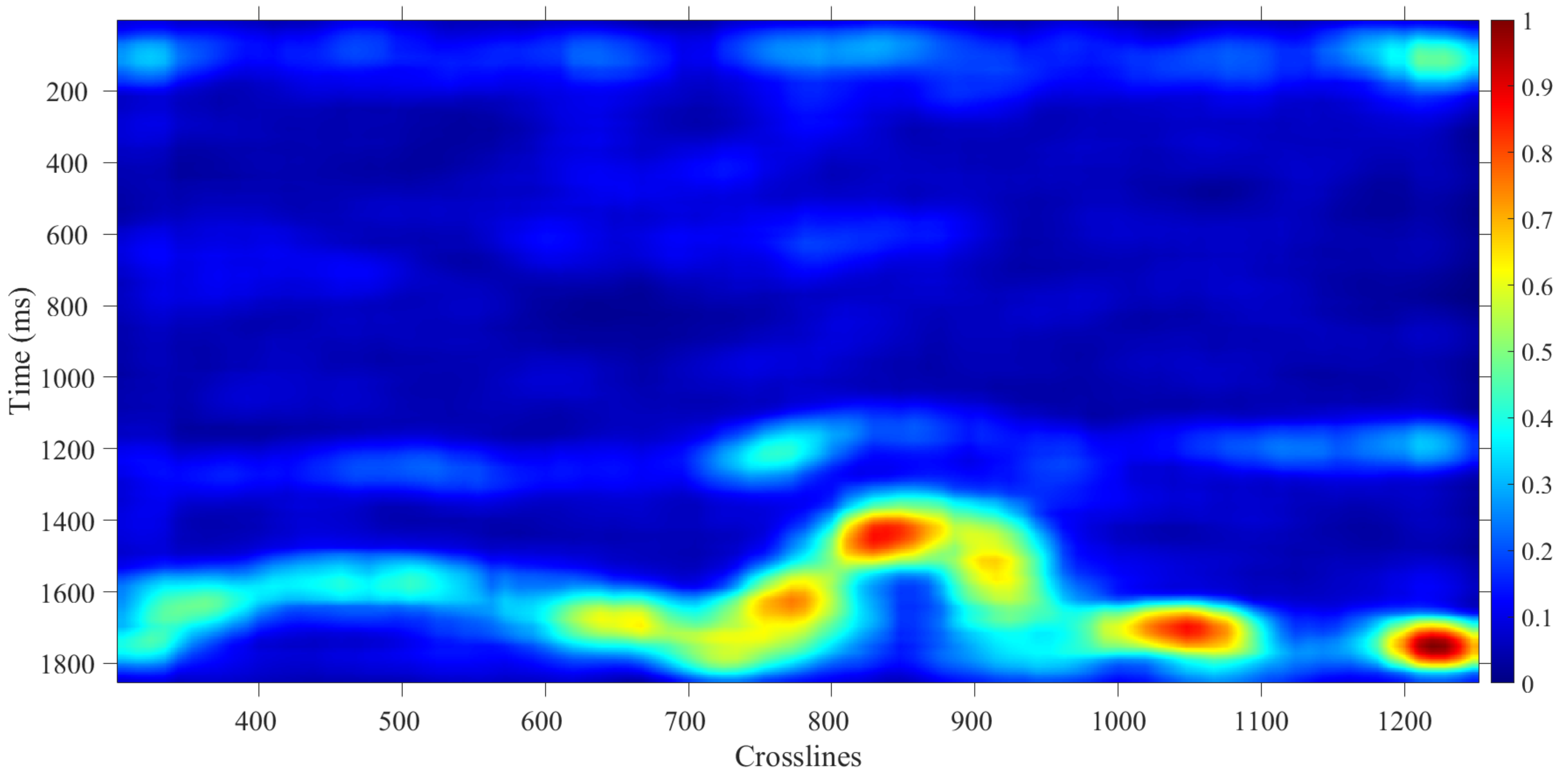}\label{fig:PQFTF3}}
  \subfigure[Achanta~\cite{achanta2008salient}]{\includegraphics[width=.325\textwidth]{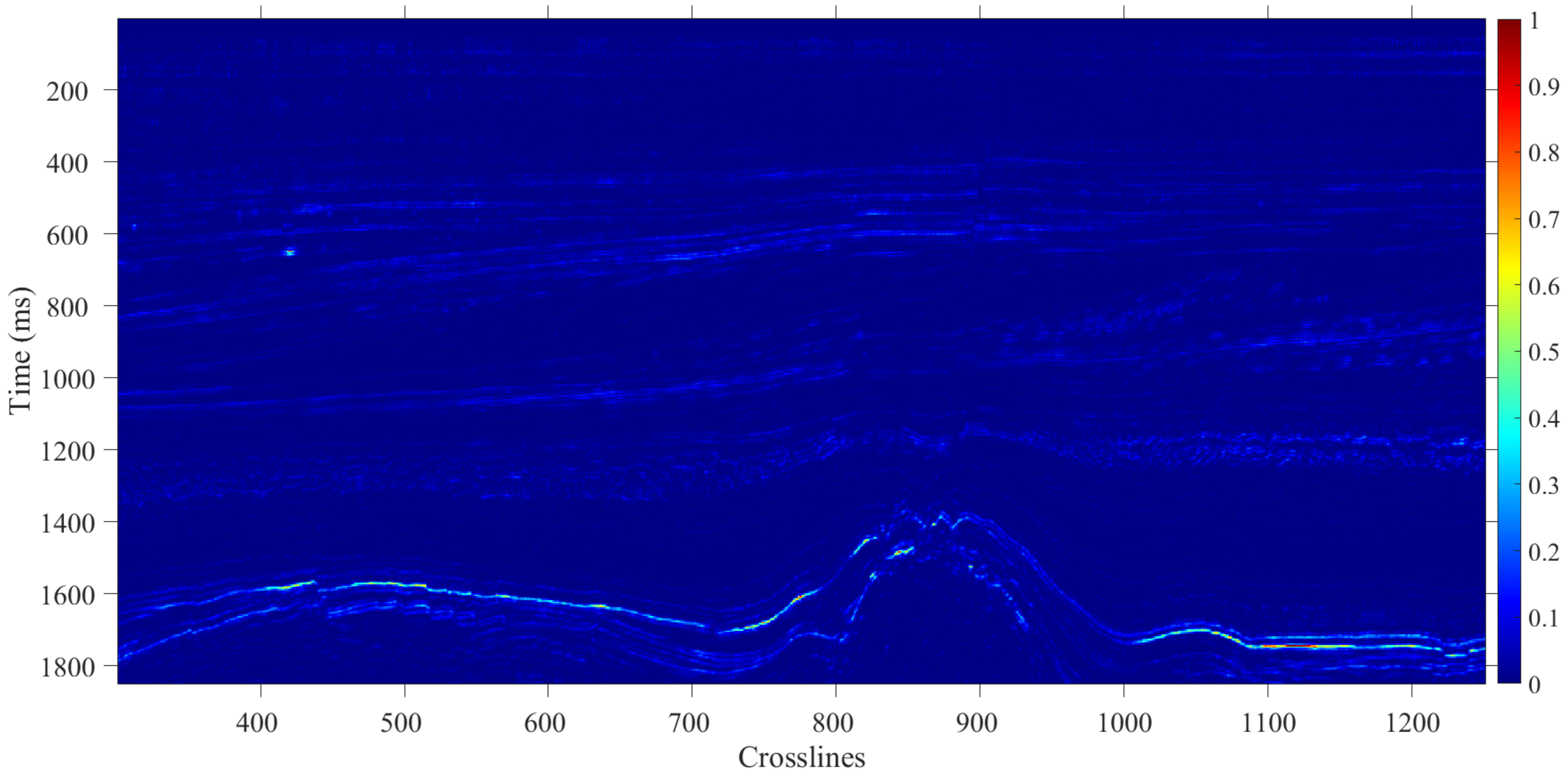}\label{fig:ICVSF3}}
  \subfigure[Fang~\cite{fang2014video}]{\includegraphics[width=.325\textwidth]{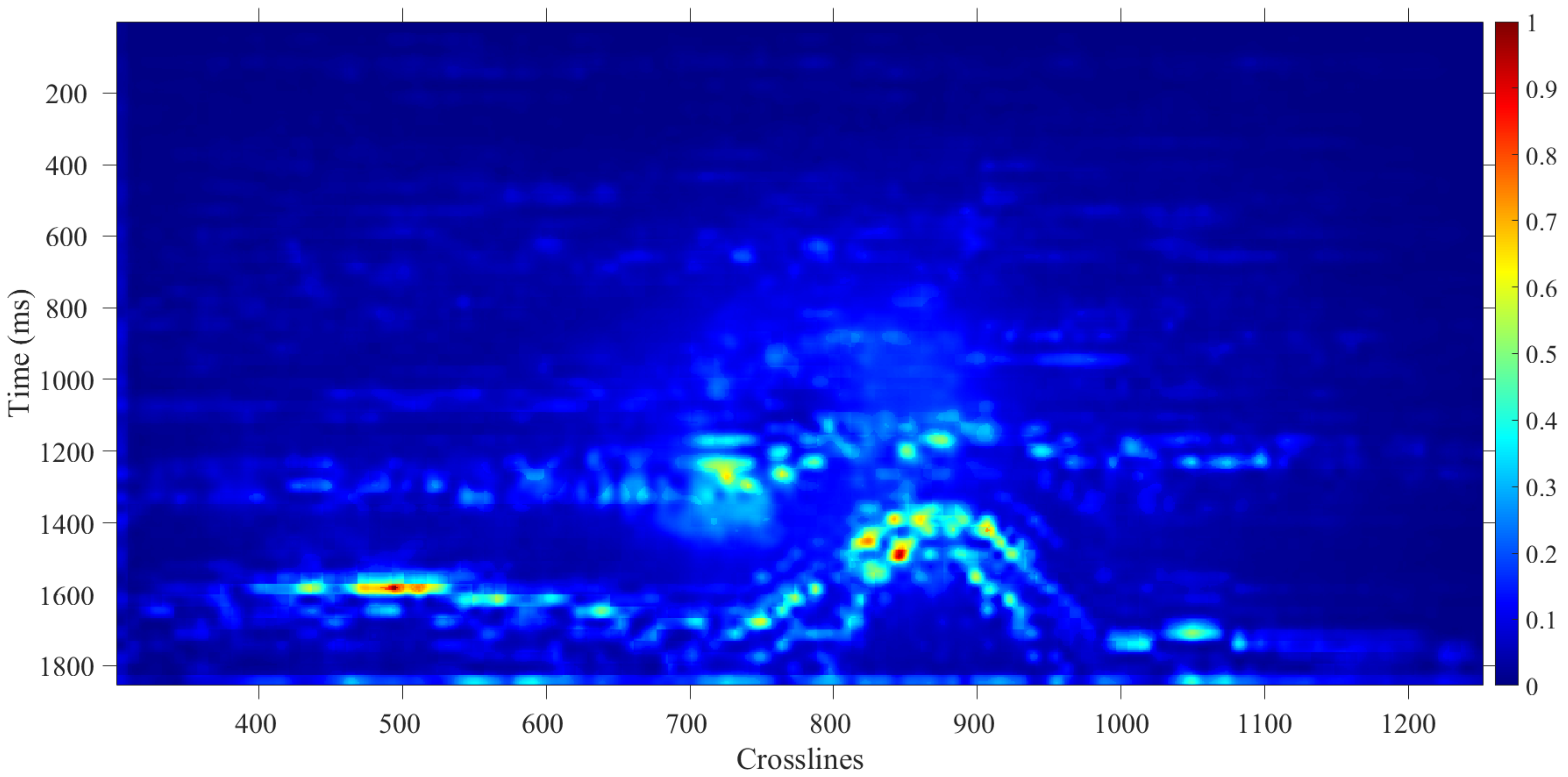}\label{fig:STUWF3}}
  \subfigure[Long~\cite{Long2015}]{\includegraphics[width=.325\textwidth]{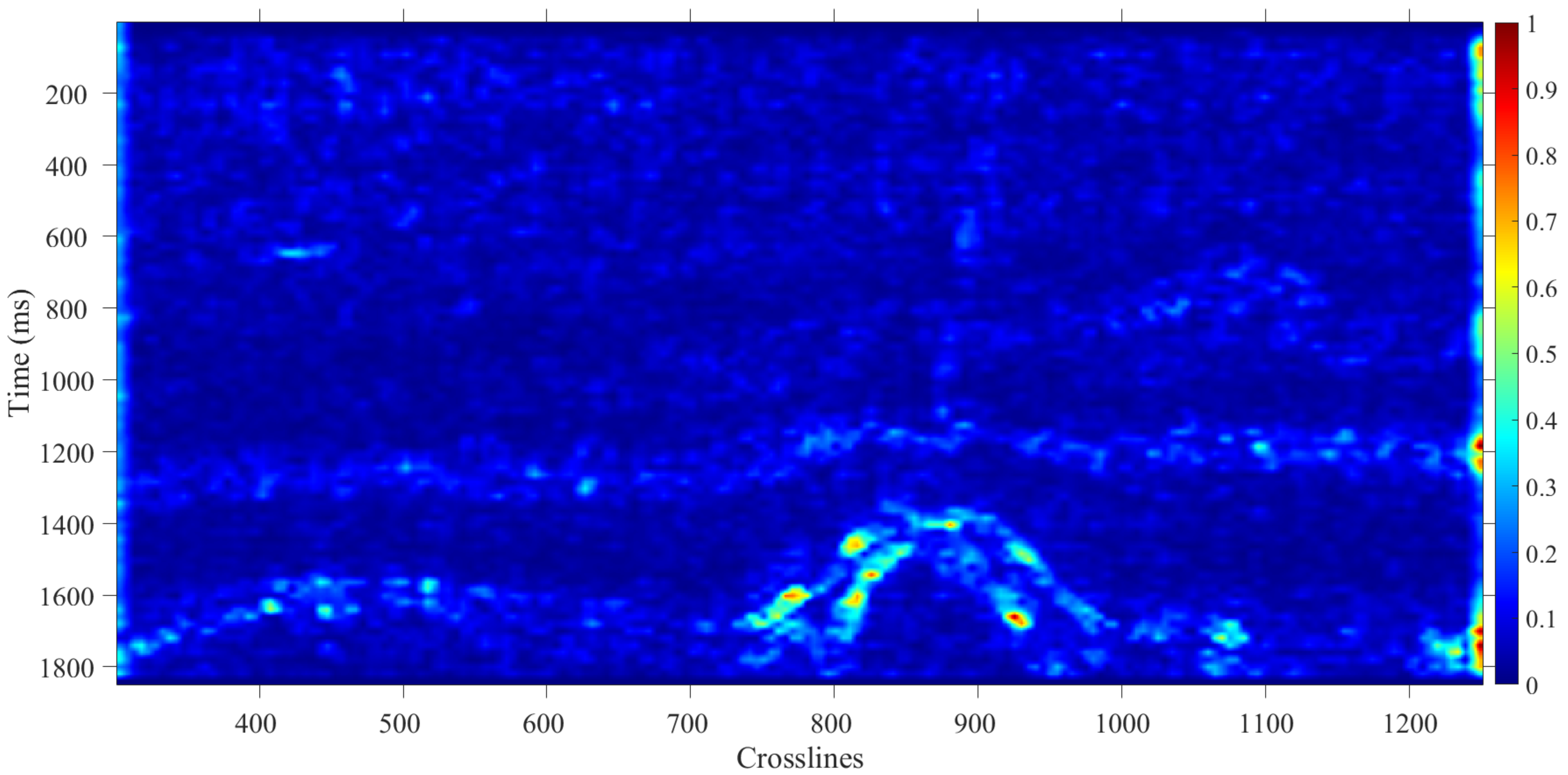}\label{fig:LongF3}}
  \subfigure[Proposed (Equal weights)]{\includegraphics[width=.325\textwidth]{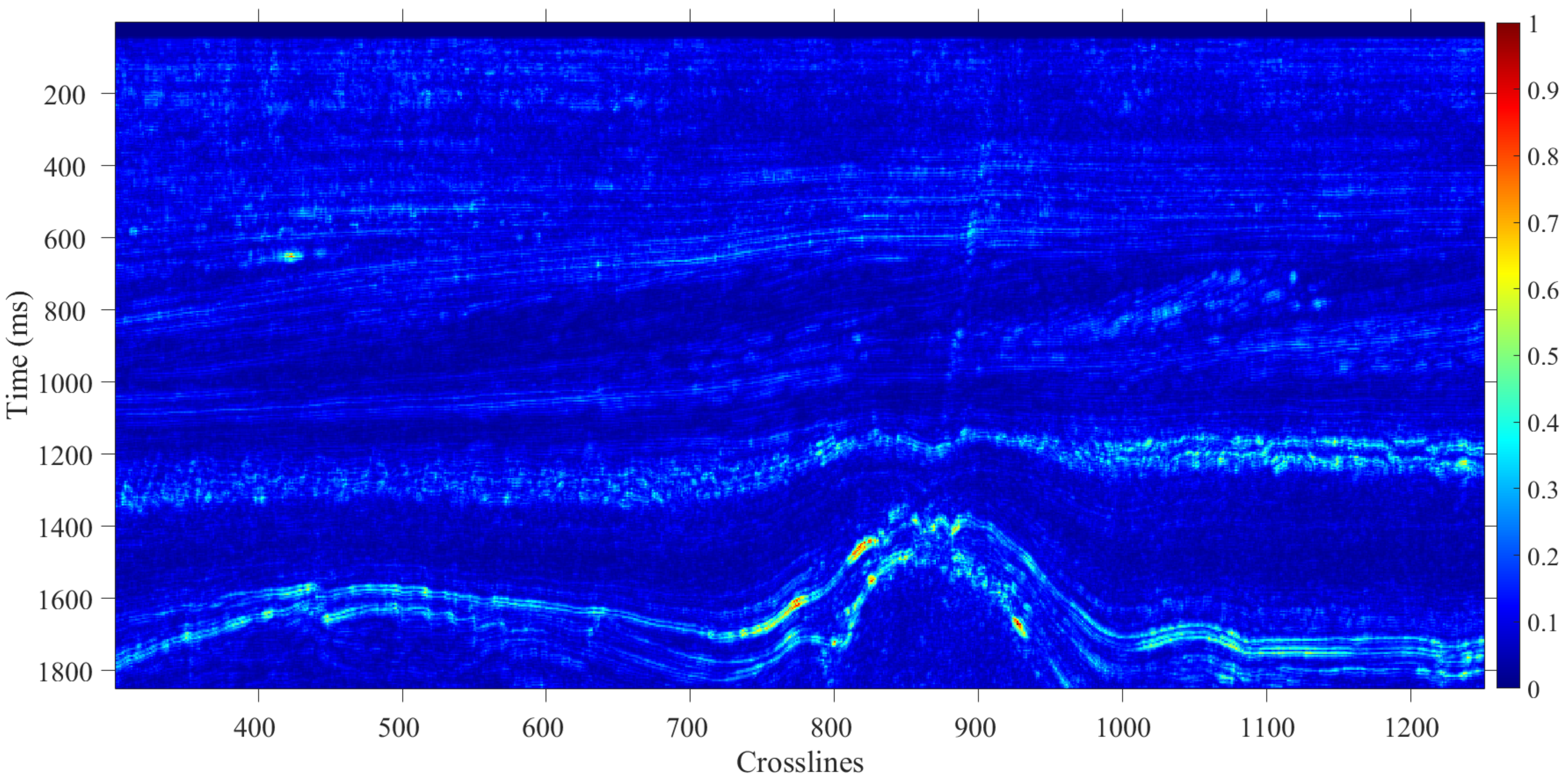}\label{fig:ProposedF3Equal}}
  \subfigure[Proposed (Adapted to highlight faults)]{\includegraphics[width=.325\textwidth]{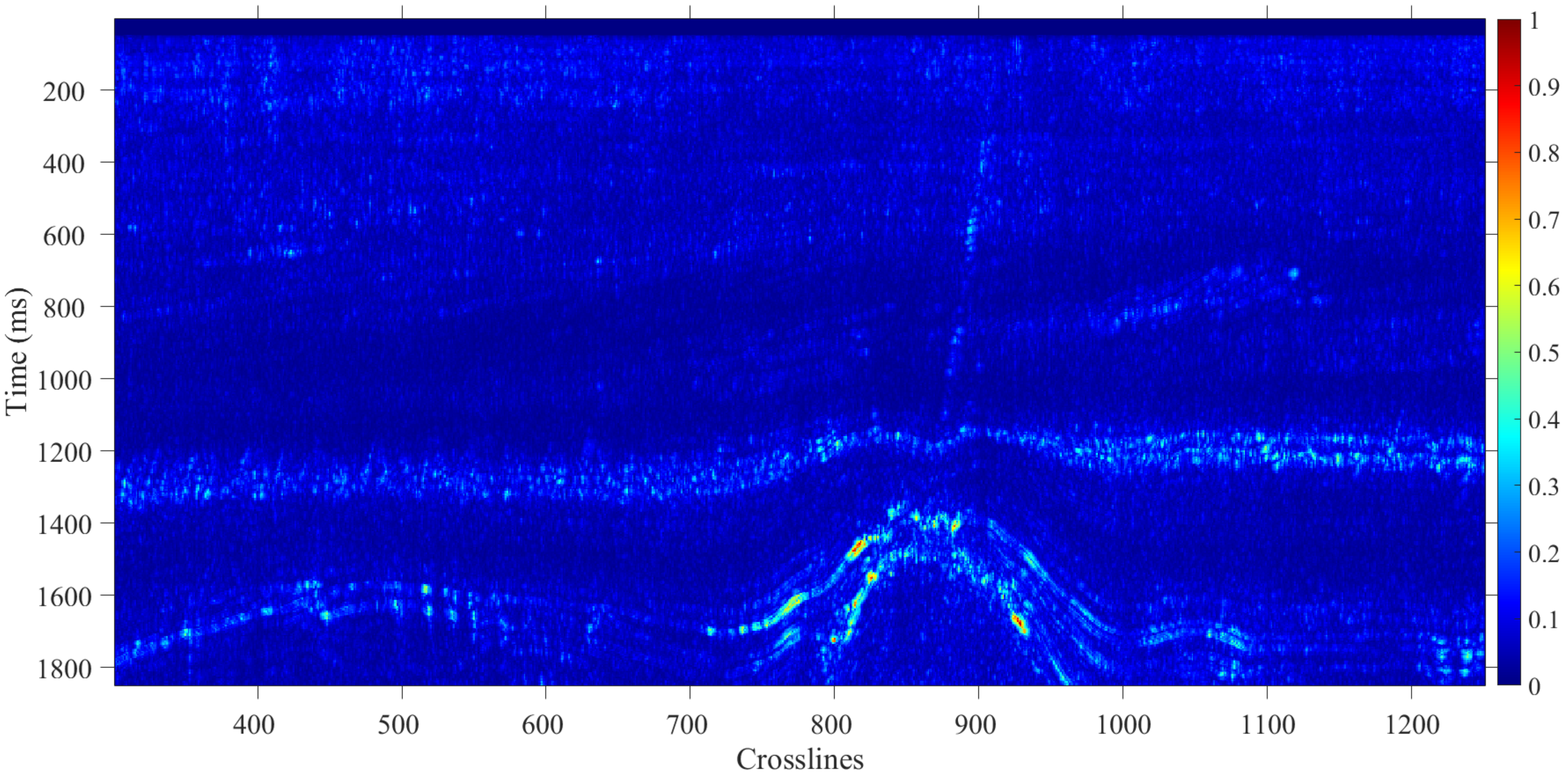}\label{fig:ProposedF3Time}}
  \vspace{-0.3cm}
  \caption{Comparison of the proposed algorithm against other algorithms on a typical seismic section within F3 block. We note that the major structures (as labeled in the original seismic image) have all been detected in the saliency maps from the proposed algorithm.}
  \label{fig:SaliencyMapsF3}
\end{figure*}

A seismic inline section (i.e., a cross section of the 3D data volume along the y-direction) from the F3 dataset containing multiple seismic structures is shown in Fig.~\ref{fig:SaliencyMapsF3}a. The results of various comparable state-of-the-art image and video saliency detection algorithms presented in \cite{zhang2008sun}, \cite{hou2007saliency}, \cite{guo2010novel}, \cite{achanta2008salient}, \cite{fang2014video}, and \cite{Long2015} are shown in Fig.~\ref{fig:SaliencyMapsF3}b-g, respectively. The saliency map from the proposed method with equal weights $W_m, m \in \{t,x,y\}$ 
is shown in Fig.~\ref{fig:SaliencyMapsF3}h, whereas the saliency map adapted to highlight faults along the time direction (i.e., the vertical direction) is displayed in Fig.~\ref{fig:SaliencyMapsF3}i.

\begin{figure*}[!ht]
  \centering
  \subfigure[Original]{\includegraphics[width=.095\textwidth]{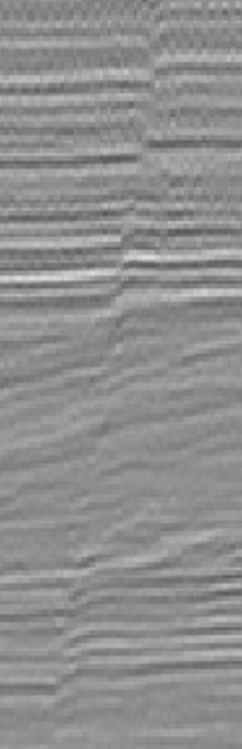}\label{fig:SS245F3_ZA}}
  \subfigure[Fang~\cite{fang2014video}]{\includegraphics[width=.1\textwidth]{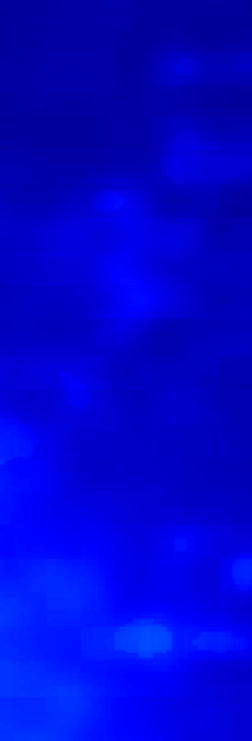}\label{fig:STUWF3_ZA}}
  \subfigure[Long~\cite{Long2015}]{\includegraphics[width=.099\textwidth]{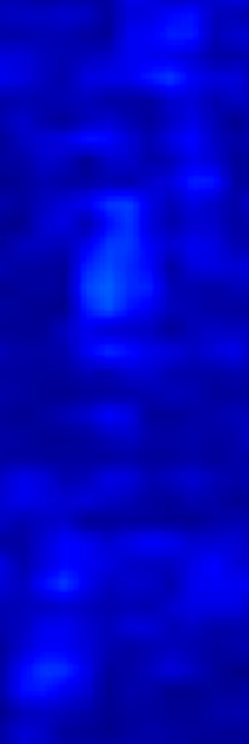}\label{fig:LongF3_ZA}}
  \subfigure[EW]{\includegraphics[width=.099\textwidth]{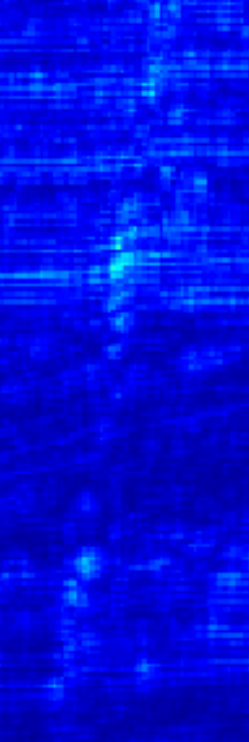}\label{fig:ProposedF3Equal_ZA}}
  \subfigure[AD]{\includegraphics[width=.098\textwidth]{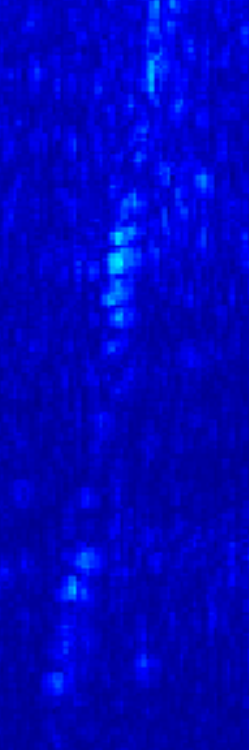}\label{fig:ProposedF3Time_ZA}}
  \vspace{-0.3cm}
  \caption{Comparison of saliency maps by zooming in on the major fault extending along the vertical direction, visible in the middle of the original seismic section within F3 block through the horizontal discontinuities. Here, EW means the proposed with equal weights, and AD means the proposed being adapted to highlight faults. It is obvious that the two zoom-in saliency maps freom the proposed algorithm (EW and AD) are capable of highlighting the fault, with the AD version presenting the best result, while the other two are not.}
  \label{fig:SaliencyMapsF3_ZA}
\end{figure*}

As observed in the resulting saliency maps, the proposed algorithm effectively highlights the salient structures from the seismic imaging data as compared to other algorithms. Specifically, Fig.~\ref{fig:SaliencyMapsF3}h shows horizons, chaotic regions, salt-dome boundary, and faults in the given seismic inline section, all at once, while most other algorithms can only highlight one or two structures, typically with inferior visual appearance showing either more artifacts or blurred boundaries. Figure~\ref{fig:SaliencyMapsF3}i highlights not only a major fault in the middle dissecting a salt dome but also a number of smaller faults at the bottom left side of the seismic section. While majority of algorithms fail to detect even the major fault in the center of the seismic section, the proposed algorithm highlights such changes successfully because it accounts for the spectral variations along all three dimensions of the seismic volume. In addition, the proposed saliency detection algorithm accentuates a sigmoidal structure, which is not distinctively detected by other algorithms.

\subsection{GSB, New Zealand}
\begin{figure*}[htp]
  \centering
    \subfigure[Saliency along time dimension, $\boldsymbol{S}_t$]{\includegraphics[width=.49\textwidth]{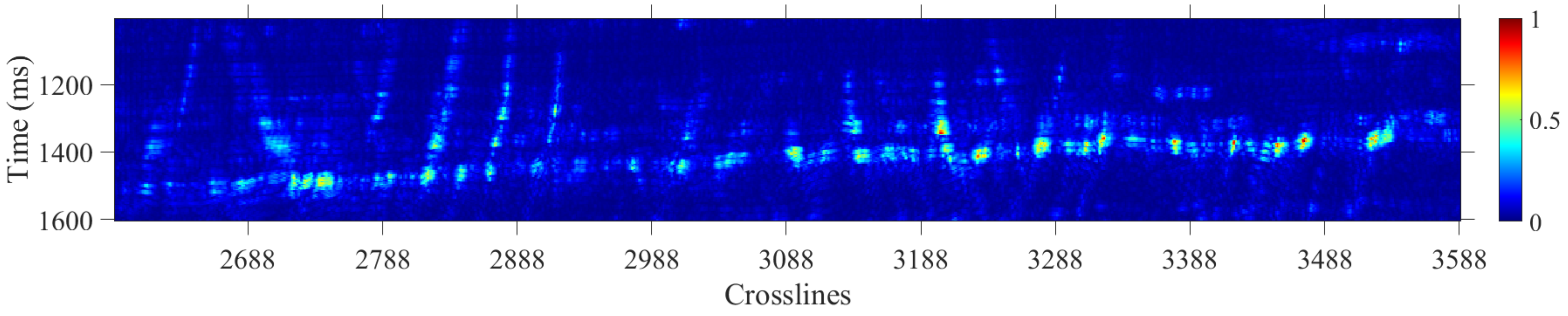}\label{fig:SS200_St}}
    \subfigure[Saliency along crossline dimension, $\boldsymbol{S}_x$]{\includegraphics[width=.49\textwidth]{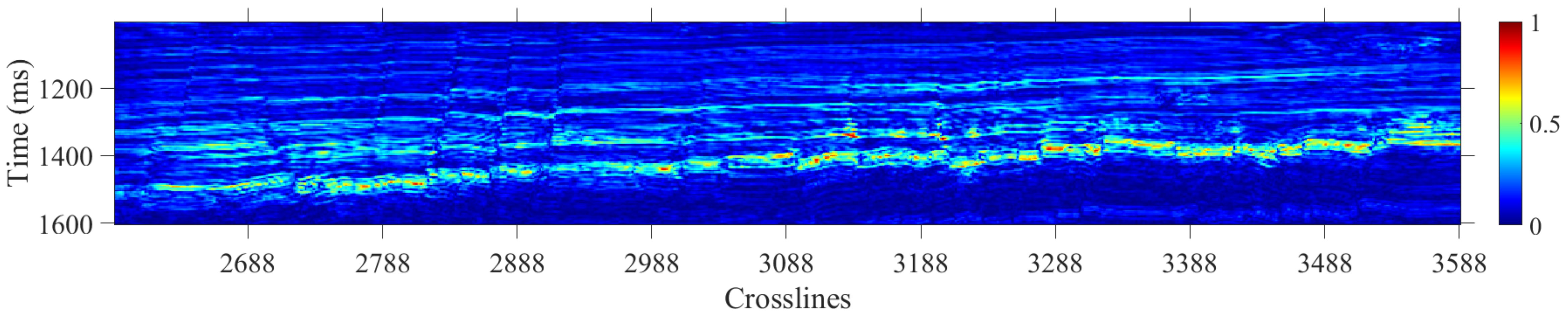}\label{fig:SS200_Sx}}
    \subfigure[Saliency along inline dimension, $\boldsymbol{S}_y$]{\includegraphics[width=.49\textwidth]{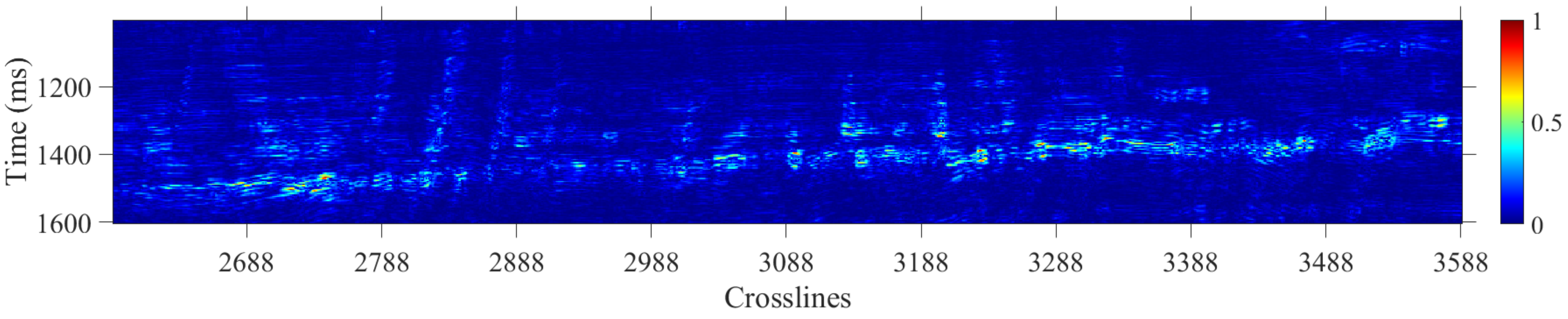}\label{fig:SS200_Sy}}
    \subfigure[Combined saliency map, $\boldsymbol{S}$]{\includegraphics[width=.49\textwidth]{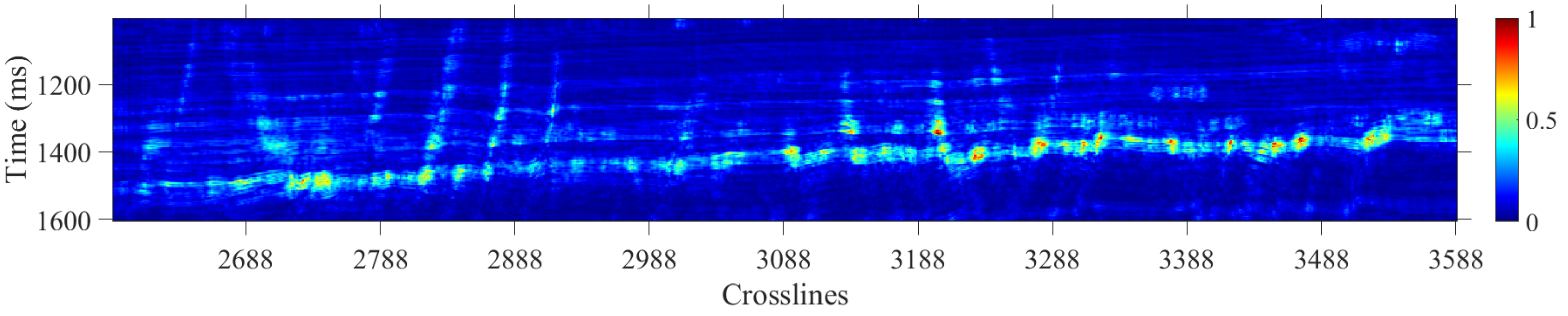}\label{fig:SS200_S}}
\caption{Example saliency maps obtained after spectral decomposition and DCS comparisons, using the GSB dataset.}
\label{fig:SaliencyProjectionsGSB}
\end{figure*}

\begin{figure*}[h]
  \centering
    \subfigure[A typical seismic time section]{\includegraphics[width=.42\textwidth]{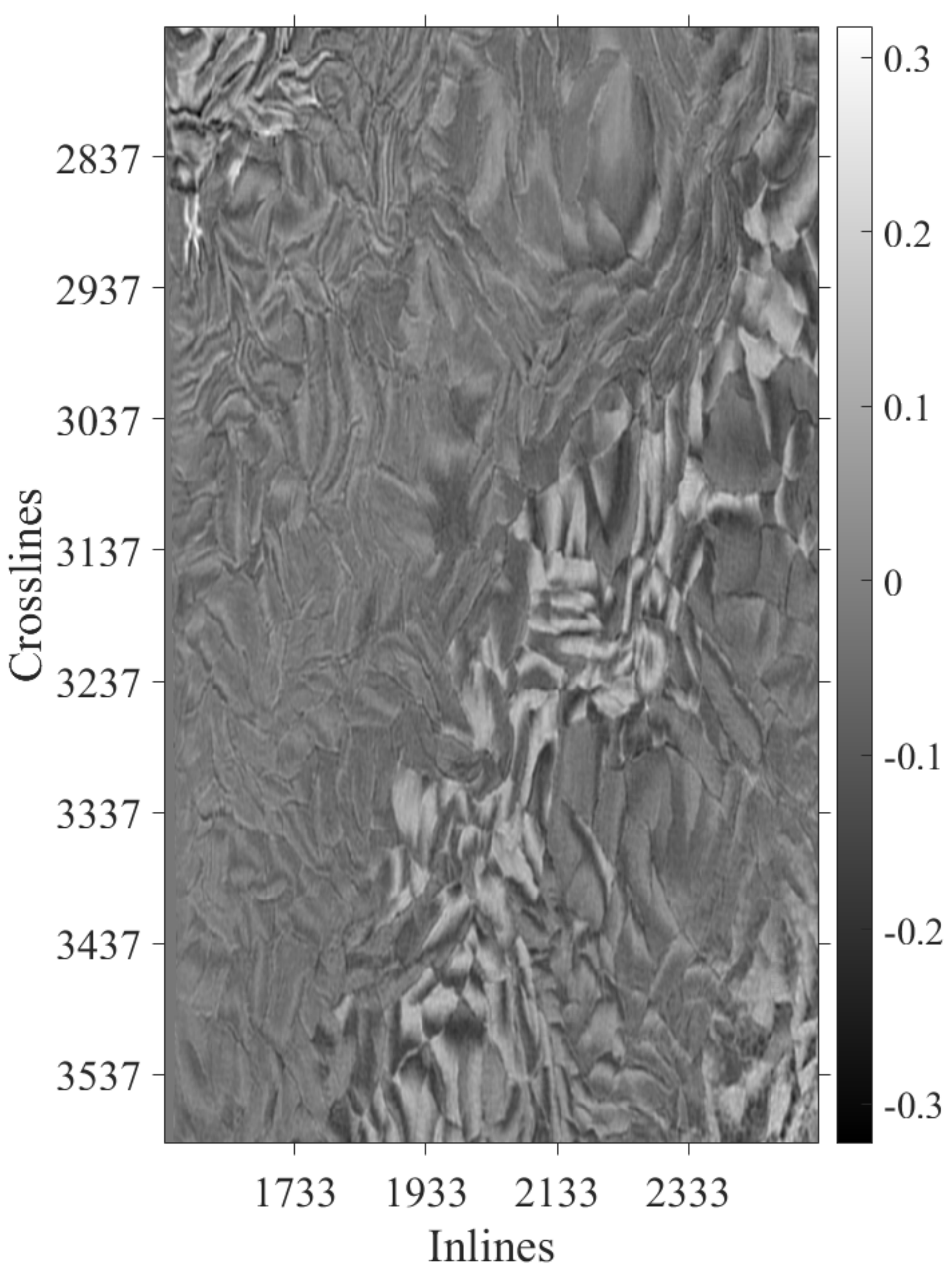}\label{fig:SS140_GSB_Time_1556ms}}
    \hspace{1cm}
    \subfigure[Saliency map]{\includegraphics[width=.42\textwidth]{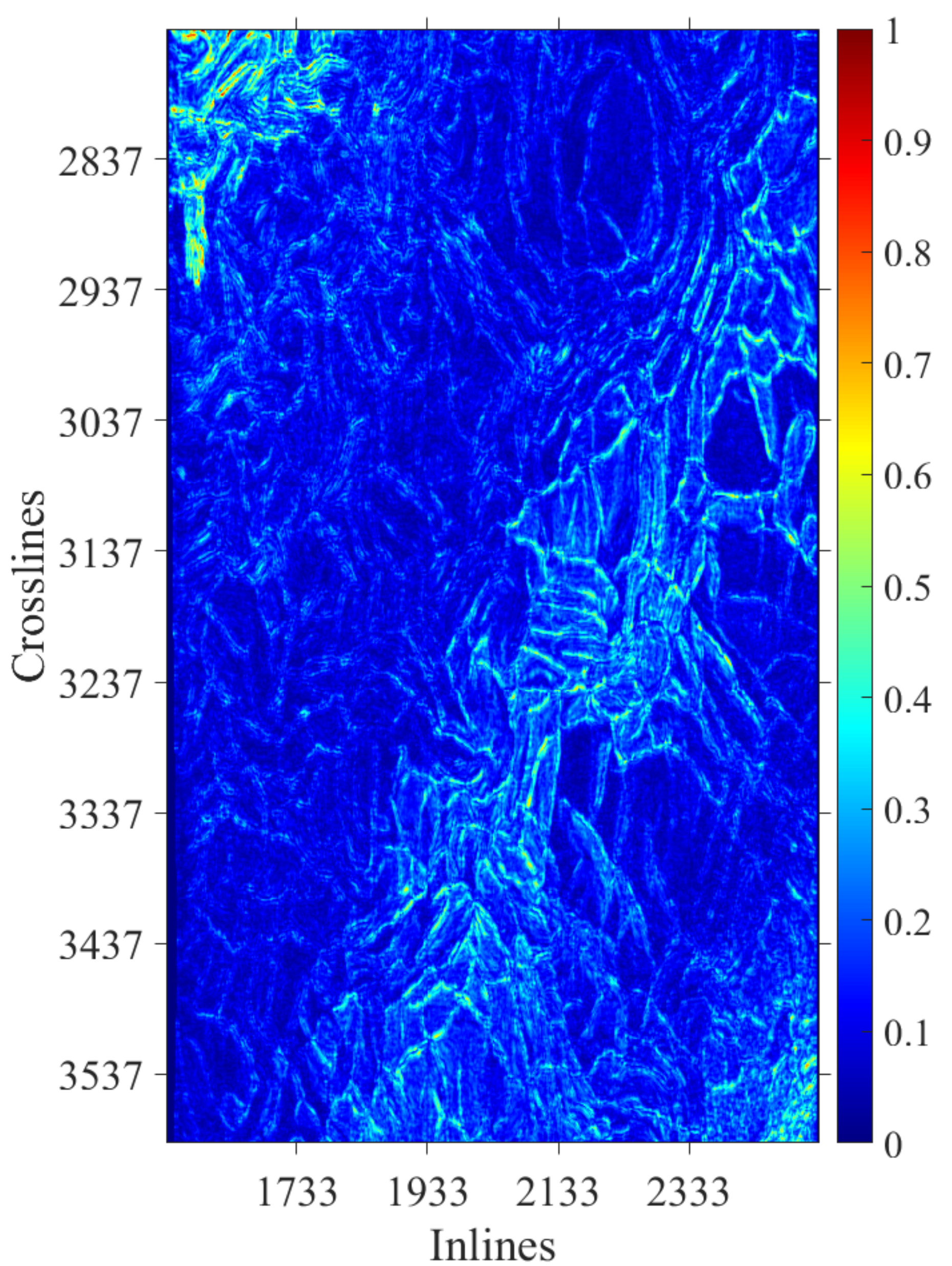}\label{fig:SS140_SalMap_GSB_SS140_Time_1556ms}}
\caption{A seismic time section within GSB dataset at 1556ms and its saliency map, which clearly highlights the complex network of faults.}
\label{fig:TimeSectionsGSB}
\end{figure*}

\begin{figure*}[htp]
  \centering
  \subfigure[A typical seismic inline section]{\includegraphics[width=.49\textwidth]{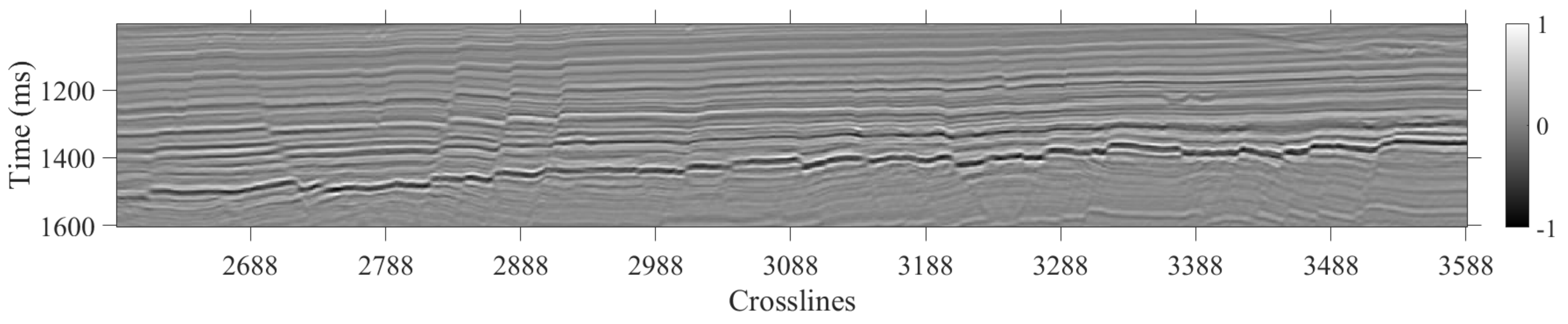}\label{fig:SS200GSB}}
  \subfigure[Ground Truth]{\includegraphics[width=.49\textwidth]{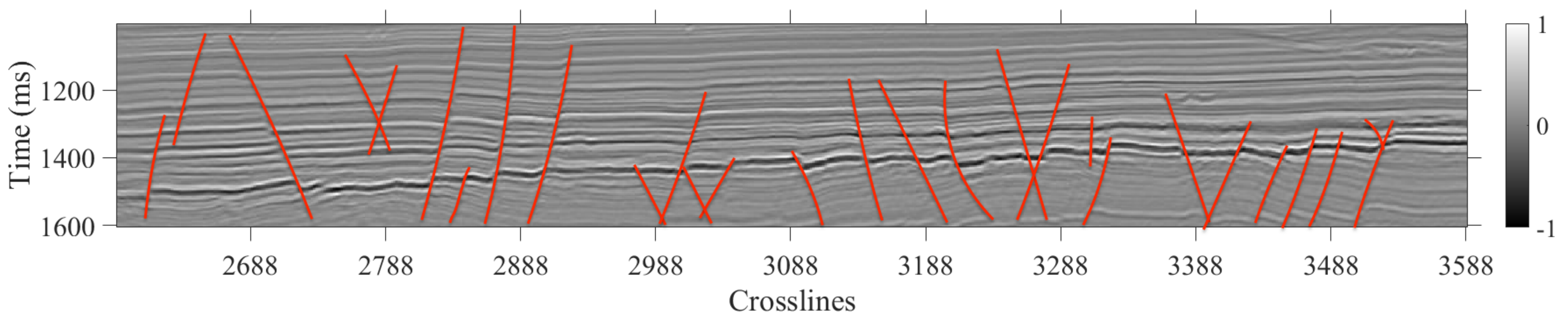}\label{fig:SS200_GTGSB}}
  \subfigure[SUN~\cite{zhang2008sun}]{\includegraphics[width=.49\textwidth]{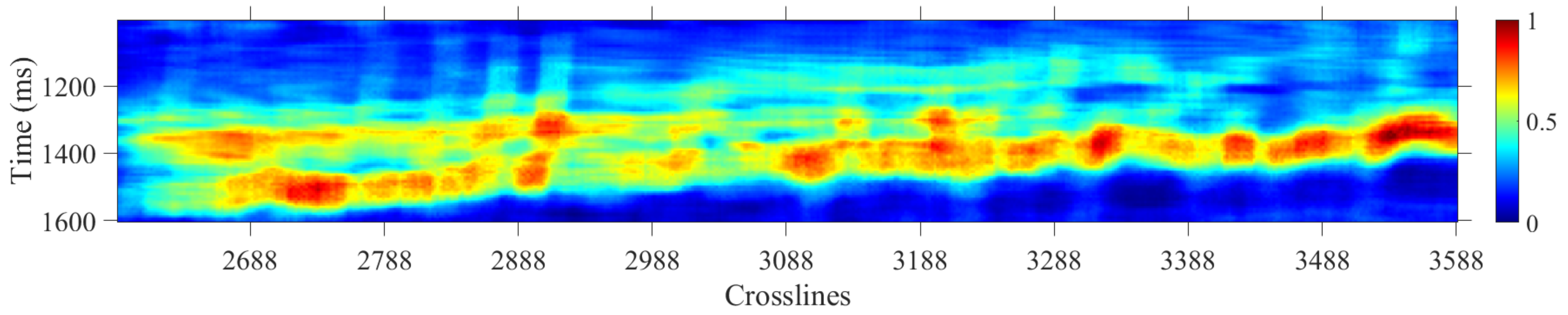}\label{fig:ISGSB}}
  \subfigure[SR~\cite{hou2007saliency}]{\includegraphics[width=.49\textwidth]{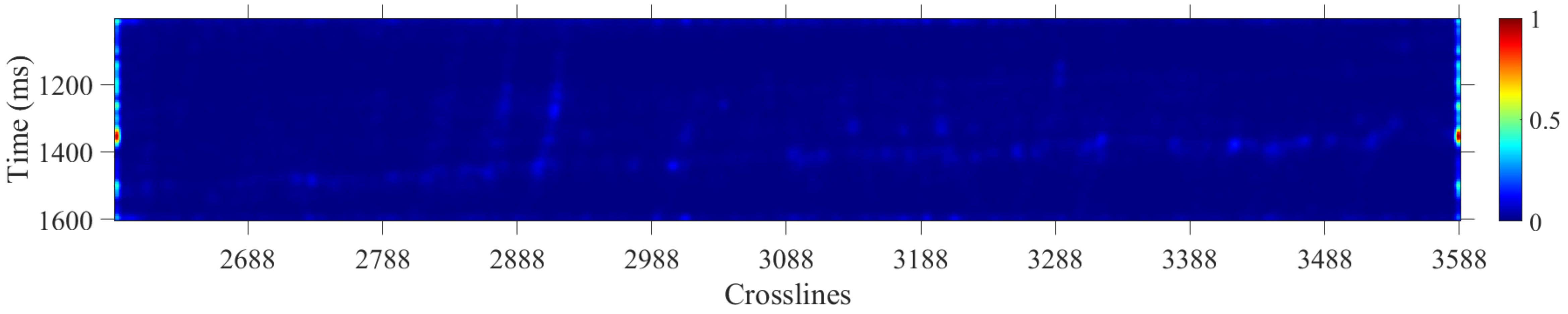}\label{fig:SResGSB}}
  \subfigure[PQFT~\cite{guo2010novel}]{\includegraphics[width=.49\textwidth]{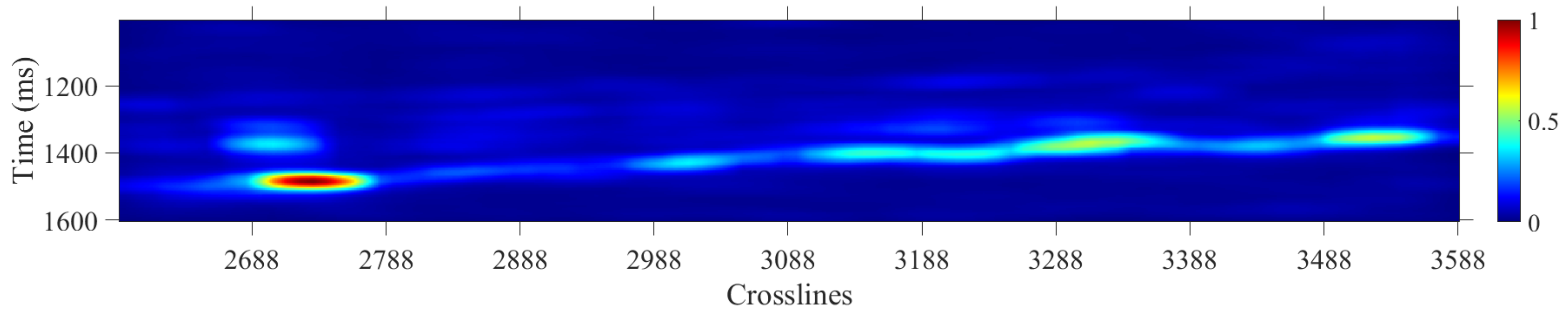}\label{fig:PQFTGSB}}
  \subfigure[Achanta~\cite{achanta2008salient}]{\includegraphics[width=.49\textwidth]{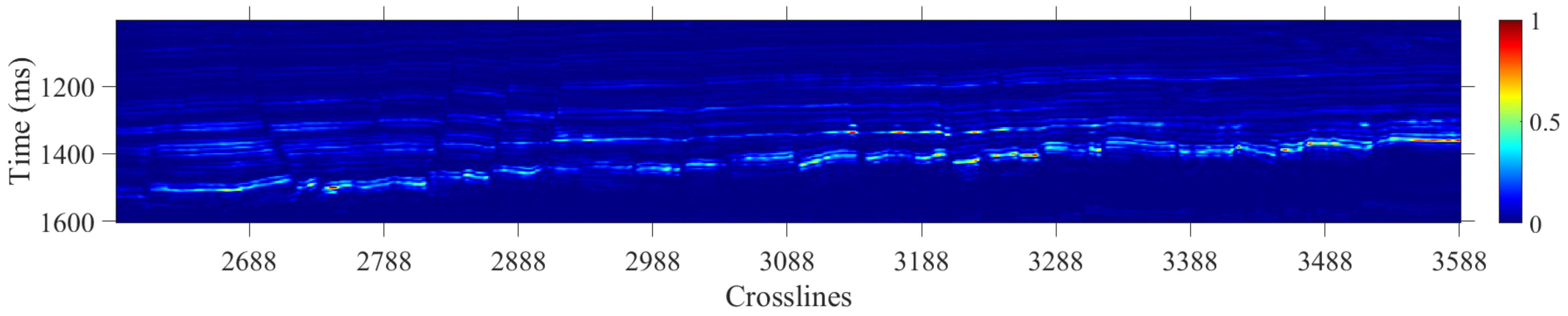}\label{fig:ICVSGSB}}
  \subfigure[Fang~\cite{fang2014video}]{\includegraphics[width=.49\textwidth]{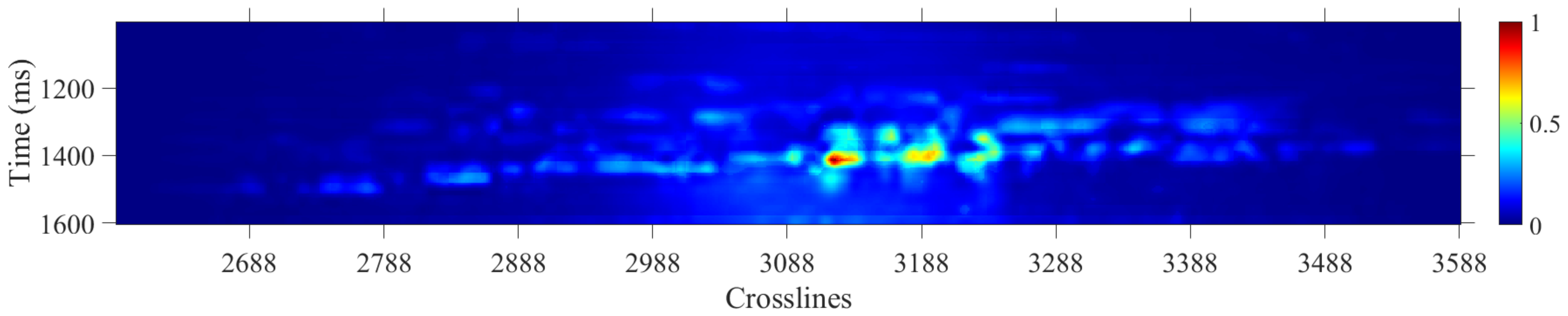}\label{fig:STUWGSB}}
  \subfigure[Long~\cite{Long2015}]{\includegraphics[width=.49\textwidth]{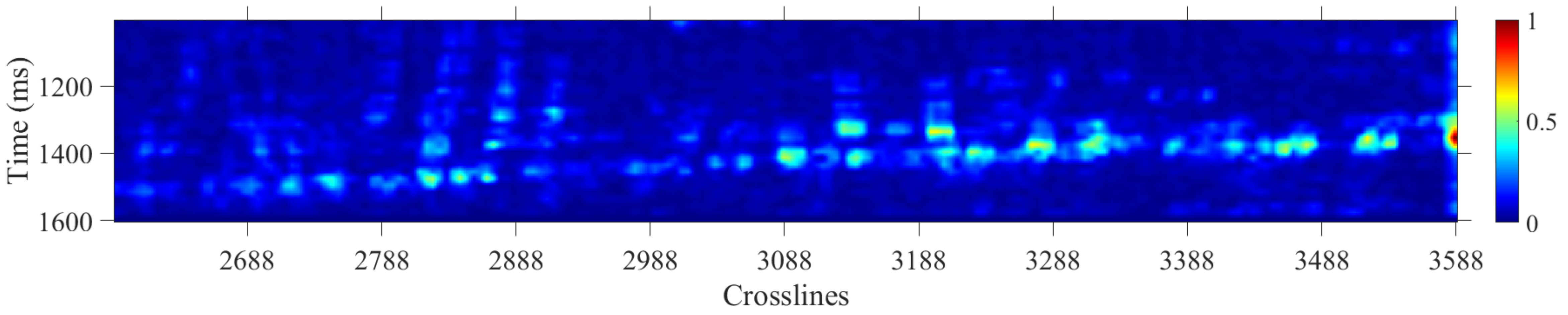}\label{fig:LongGSB}}
  \subfigure[QSS~\cite{schauerte2012quaternion}]{\includegraphics[width=.49\textwidth]{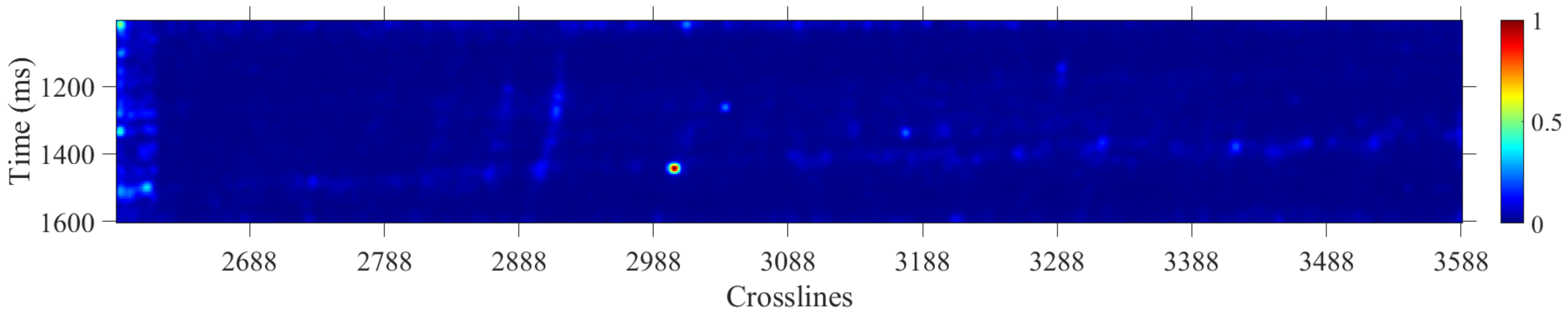}\label{fig:SSGSB}}
  \subfigure[Seo~\cite{seo2009static}]{\includegraphics[width=.49\textwidth]{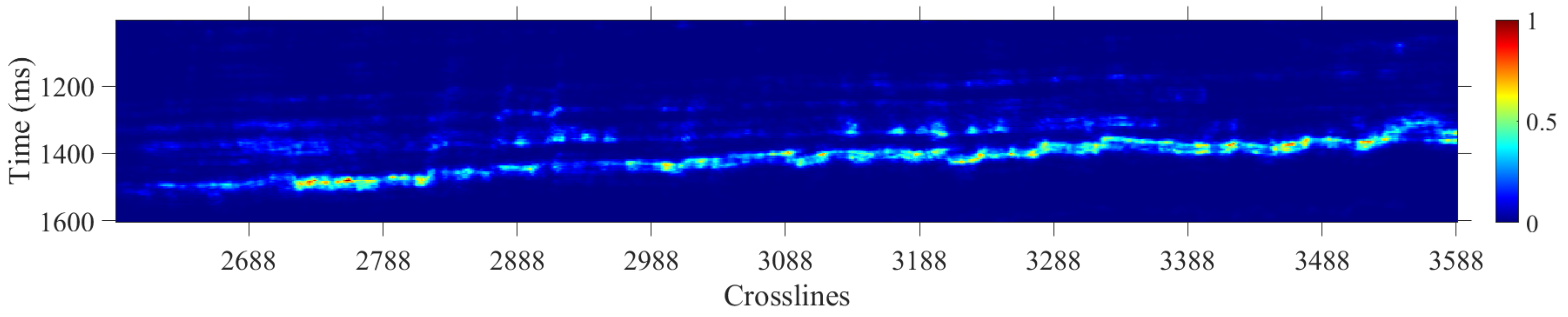}\label{fig:SRGSB}}
  \subfigure[GTC~\cite{yazeedGTC}]{\includegraphics[width=.49\textwidth]{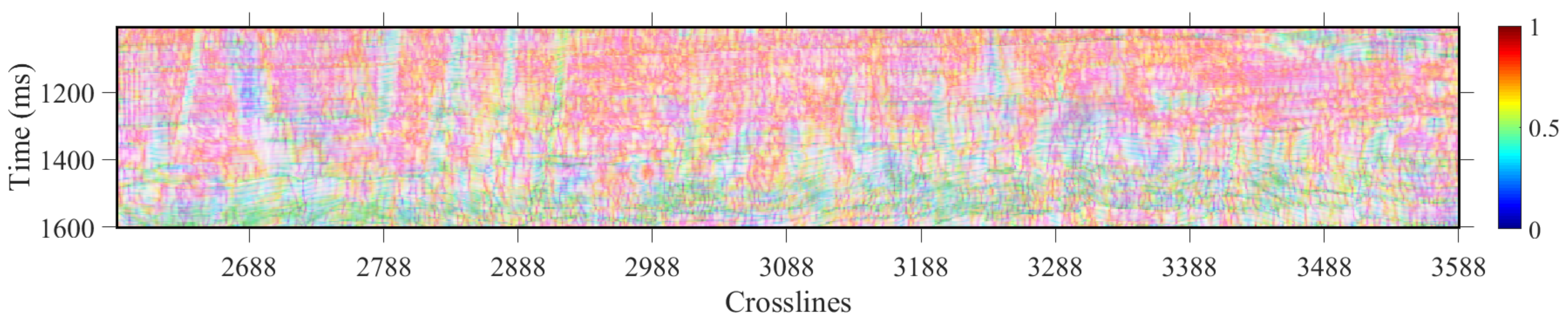}\label{fig:GTCGSB}}
  \subfigure[C-Cube\cite{CoherenceCube}]{\includegraphics[width=.49\textwidth]{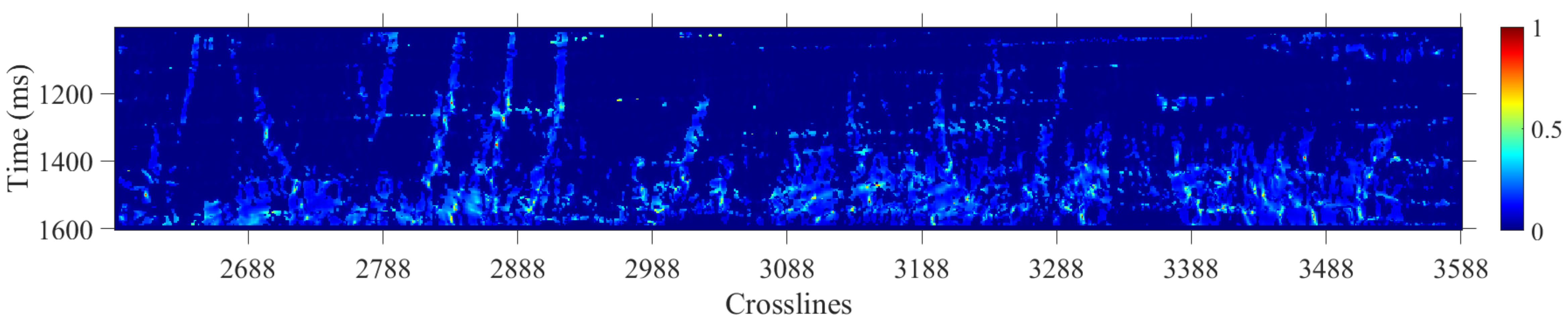}\label{fig:DisconGSB}}
  \subfigure[Proposed Method (Equal weights)]{\includegraphics[width=.49\textwidth]{Figs/GSB/SS200_S}\label{fig:ProposedEqGSB}}
  \subfigure[Proposed Method (Larger $W_t$)]{\includegraphics[width=.49\textwidth]{Figs/GSB/SS200_St}\label{fig:ProposedT}}
  \caption{Comparison of the proposed algorithm against various saliency detection algorithms, as well as non-saliency-based algorithms, on detecting a complex network of faults in a typical seismic inline section within GSB dataset. The proposed algorithm generated satisfactory results, while most others failed to do the job.}
  \label{fig:SaliencyMapsGSB}
\end{figure*}

The dataset from Great South Basin (GSB), New Zealand, comprises of complex networks of faults inter-winded and dwindled together. While some faults can be observed easily, others are very subtle and not easily discernible. To examine the performance of our proposed algorithm on this challenging dataset, first of all, we show the saliency maps $S_m, m \in \{t,x,y\}$ obtained for a typical seismic inline section (Fig.~\ref{fig:SaliencyMapsGSB}a) in Fig.~\ref{fig:SaliencyProjectionsGSB}a-c and a consolidated saliency map obtained by combining these saliency maps using equal weights in Fig.~\ref{fig:SaliencyProjectionsGSB}d. It can be observed clearly that each map depicts the spectral variations and relative motion of structures along their respective dimensions. Furthermore, the combined saliency map highlights all changes in a unified manner without any manual tuning. Again, if needed, it can also be adapted to highlight changes based on a priori information such as shape, orientation or structures specified in a desired map. Second, we present a time section from the GSB dataset and its saliency map in Fig.~\ref{fig:TimeSectionsGSB}a-b, respectively. As observed, the obtained saliency map effectively highlights complex fault networks that are not evident in the original seismic section. The subtle variations and overlapping faults can be easily recognized from the saliency map. 

For a comparative study, we examined a seismic inline section from GSB dataset containing multiple faults, as shown in Fig.~\ref{fig:SaliencyMapsGSB}a. The ground truth for this seismic section, with faults manually labelled by a geophysicist, is shown in Fig.~\ref{fig:SaliencyMapsGSB}b. The results of comparable state-of-the-art saliency detection algorithms presented in \cite{zhang2008sun}, \cite{hou2007saliency}, \cite{guo2010novel}, \cite{achanta2008salient}, \cite{fang2014video}, \cite{Long2015}, \cite{schauerte2012quaternion}, and \cite{seo2009static} are shown in Fig.~\ref{fig:SaliencyMapsGSB}c-j, respectively. The detection results from two non-saliency-based algorithms are also shown in Fig.~\ref{fig:SaliencyMapsGSB}k-l, respectively, of which one is based on the generalized tensor coherence~\cite{yazeedGTC} and the other based on the coherence cube~\cite{CoherenceCube}. The saliency map generated from the proposed algorithm with equal weights $W_m$ is shown in Fig.~\ref{fig:SaliencyMapsGSB}m; whereas Fig.~\ref{fig:SaliencyMapsGSB}n displays the saliency map from the proposed method adapted to highlight faults along the time dimension. 

From Fig.~\ref{fig:SaliencyMapsGSB}, it is clear that most available saliency-based algorithms in the comparative study failed to detect any or vast majority of the faults. Only two of them, that is, the ones from~\cite{Long2015} and~\cite{seo2009static}, were able to capture some of the faults, but not with a distinct appearance. In contrast, from Fig.~\ref{fig:SaliencyMapsGSB}m-n we observe that the proposed algorithm effectively highlighted all the major faults present within the seismic section. Figure~\ref{fig:SaliencyMapsGSB}n demonstrates that, by adapting to directional saliency along the time dimension, our proposed algorithm not only tuned out clutters, but also presented the faults distinctly with each fault easily distinguishable from others. As to the two non-saliency-based detection results using~\cite{yazeedGTC} and~\cite{CoherenceCube}, they contain too much noise, not able to highlight the faults adequately. 

From the studies above on the two real seismic datasets, it is evident that the proposed algorithm can effectively detect subsurface structures of diversified natures and appearances in one shot. It is also capable of capturing subtle spatial variations characteristic of some structures such as faults. The algorithm performs consistently across various structures and different datasets, generating saliency maps of excellent qualities. The amplitude of salient values detected by the proposed algorithm are significantly higher for the structures of interest as compared to other methods. In the mean time, the detected salient areas are more localized near the structures of interest. Furthermore, the resolution of the results generated from the proposed algorithm is typically much better than those from the other algorithms, which makes it advantageous for applications requiring not only fine perception but also efficient detection of subtle signatures. 

\newpage
\section{Conclusion}
\label{sec:conclusion}
In this paper, we presented a new approach to seismic interpretation applications based on visual perception and modeling of human visual system. We proposed a saliency detection algorithm based on a novel attention model for identifying subsurface structures within seismic data volumes. The proposed algorithm extracts features based on 3D-FFT local spectra and multi-dimensional spectral projections. It utilizes a novel directional center-surround attention model to compute saliency maps along different dimensions within seismic volume, and then adaptively combines them to yield a consolidated saliency map. Using the proposed algorithm, we can incorporate seismic domain knowledge by specifying directional comparisons and adapting the saliency map to desired characteristics that effectively highlight subtle signatures and structures in a geologically complex setting. Experimental results on two real seismic datasets from the North Sea, Netherlands and Great South Basin, New Zealand demonstrated the effectiveness of the proposed algorithm for detecting salient seismic structures of different natures and appearances in one shot, which differs significantly from traditional seismic interpretation algorithms. The results further demonstrated that the proposed method outperforms comparable state-of-the-art saliency detection algorithms, which are more suitable for natural images and videos, but not being adequate for seismic imaging data.

\section*{Conflict of interest}
All authors declare no conflicts of interest in this paper.

\end{document}